\documentclass{article} 
\usepackage{iclr2026_conference,times}

\usepackage{amsmath,amsfonts,bm}

\def\eqref#1{equation~\ref{#1}}
\def\Eqref#1{Equation~\ref{#1}}

\def\1{\bm{1}}

\DeclareMathAlphabet{\mathsfit}{\encodingdefault}{\sfdefault}{m}{sl}
\SetMathAlphabet{\mathsfit}{bold}{\encodingdefault}{\sfdefault}{bx}{n}

\usepackage{wrapfig}
\usepackage{hyperref}
\usepackage{url}
\usepackage{booktabs}
\usepackage{diagbox}
\usepackage{bm}
\usepackage{amsmath}

\usepackage{amssymb,amsfonts}
\usepackage[ruled,linesnumbered, vlined]{algorithm2e}
\usepackage[]{algpseudocode}                               
\algrenewcommand\textproc{\texttt}
\usepackage{caption}
\usepackage{bigstrut}
\usepackage{subfig}
\usepackage{enumitem}
\usepackage{float}
\usepackage{balance}
\usepackage{multirow}
\usepackage{ulem}
\usepackage{tabularx}
\usepackage{pdfpages}
\usepackage{graphicx} 
\usepackage{threeparttable}
\usepackage{makecell}
\usepackage{cleveref}
\usepackage{natbib}
\usepackage{amsthm}

\theoremstyle{plain}
\newtheorem{theorem}{\textbf{Theorem}}
\theoremstyle{definition}
\newtheorem{definition}{\textbf{Definition}}

\newcommand{\fakeparagraph}[1]{\vspace{1mm}\noindent\textbf{#1.}}

\iclrfinalcopy

\title{LoopServe:  An Adaptive Dual-phase LLM Inference Acceleration System for Multi-Turn Dialogues}

\author{Haoyang Li\textsuperscript{1} \  Zhanchao Xu\textsuperscript{1,3} \  Yiming Li\textsuperscript{4} \  Xuejia Chen\textsuperscript{1,3} \  Darian Li\textsuperscript{1} \  Anxin Tian\textsuperscript{2}\\
	 \textbf{Qingfa Xiao\textsuperscript{5} \  Cheng Deng\textsuperscript{6} \  Jun Wang\textsuperscript{7} \  Qing Li\textsuperscript{1} \  Lei Chen\textsuperscript{5} \  Mingxuan Yuan\textsuperscript{4}}
	\\[1.5mm]
	\small \textsuperscript{1}{PolyU} ~\  \textsuperscript{2}{HKUST} ~\  \textsuperscript{3}{HUST} ~\  \textsuperscript{4}{Huawei Noah’s Ark Lab}  ~\  \textsuperscript{5}{HKUST(GZ)} ~\  \textsuperscript{6}{Edin} ~\  \textsuperscript{7}{UCL} \\
	\texttt{haoyang-comp.li@polyu.edu.hk}\\
}

\begin{document}

\maketitle

\begin{abstract}
	Multi-turn dialogues are essential in many real-world applications of large language models, such as chatbots and virtual assistants. As conversation histories become longer, existing large language models face increasing computational and memory challenges, which hinder their ability to provide efficient and responsive interactions. Most current acceleration methods either compress the context or optimize key value caching, but they often rely on fixed or position-based heuristics that do not adapt well to the dynamic and unpredictable patterns found in actual multi-turn conversations. As a result, these models   cannot accurately identify and prioritize the most relevant context, leading to degraded response quality. In this paper, we present LoopServe, an adaptive dual-phase inference acceleration framework for large language models in multi-turn dialogues. LoopServe introduces two main innovations. 	First, it performs online sparsification during the prefilling phase by dynamically selecting the most important parts of the attention matrix for each new input.  Second, it uses progressive key value compression during decoding by adaptively maintaining a relevant and efficient cache based on the most recently generated output tokens. We also propose a new benchmark with eleven multi-turn datasets that reflect realistic query positions and conversational dependencies. Extensive experiments demonstrate that LoopServe consistently achieves superior effectiveness compared to existing baselines and significantly accelerates LLM inference across a wide range of long-context dialogue tasks.
\end{abstract}

\section{Introduction}\label{sec:intro}
 Multi-turn dialogues are at the core of numerous real-world applications, from customer service chatbots to virtual assistants and collaborative agents. 
These scenarios demand that large language models (LLMs)~\citep{hadi2023survey,deng2025plm, journals/corr/abs-2412-19442,zhou2024db,journals/pvldb/RenenSK24}
 not only generate coherent responses but also maintain contextual consistency across lengthy, evolving conversations. As the number of dialogue turns increases,
  so does the computational workload. 
For instance, processing a multi-turn dialogue comprising 10,000 tokens with Llama-3.1-70B~\citep{2024llama3herdmodels} can demand trillions of floating-point operations (FLOPs), quickly challenging the limits of real-time inference.
  Despite the remarkable progress of LLMs such as GPT~\citep{NEURIPS2020_1457c0d6,radford2018improving,radford2019language}, 
  Llama~\citep{2024llama3herdmodels,touvron2023llama}, 
  and DeepSeek~\citep{deepseekv3,deepseekr1}, 
  their inefficiency in handling multi-turn dialogues remains largely unaddressed.

In a typical multi-turn setting, each new user turn expands the conversation history, requiring the LLM to process ever-growing input sequences. 
The model’s self-attention mechanism, which lies at the heart of the Transformer~\citep{vaswani2017attention}, needs to compute pairwise attention scores between every pair of tokens in the input. 
Specifically, given an input sequence of length $n$, an LLM with $P$ parameters and a hidden dimension $d$, the time complexity of generating $m$ tokens is $\mathcal{O}(m((n+m)^2d + P))$.
For instance, in a 3-turn conversation with 5000 tokens per turn, the effective context length for the model reaches 15,000 tokens, resulting in quadratic growth in both computational cost and memory usage. This compounding effect of context accumulation makes real-time, cost-efficient inference increasingly difficult as the conversation progresses.
Different from single-turn tasks,
the context in multi-turn dialogues accumulates dynamically and queries may appear at the beginning, middle, or end of the input, causing attention patterns to shift unpredictably. The resulting attention matrices not only grow with each turn but also exhibit highly dynamic and input-dependent sparsity, exacerbating the inefficiency of current inference methods.

Recent research proposes accelerating LLM inference by reducing the computational burden of attention weight calculations during both the prefilling and decoding stages.
In the prefilling stage, where the attention matrix is computed for all token pairs, methods~\citep{jiang2024minference,lv2024critiprefill,lai2025flexprefill}, such as Minference~\citep{jiang2024minference}, use fixed pattern to sparsify the attention matrix to reduce quadratic computation.
During decoding, KV caches store precomputed \textsf{Key} and \textsf{Value} vectors to reduce redundant computation. Methods like H2O~\citep{zhang2023h2o}, SnapKV~\citep{li2024snapkv}, and AdaKV~\citep{feng2024ada} cache tokens selected based on tokens at the end of the query. However, these approaches rely on static or position-based heuristics and cannot adapt to the dynamic, input-dependent patterns of real multi-turn dialogues.

Moreover, current evaluation benchmarks~\citep{journals/corr/abs-2412-19442,li2025exposingnumeracygapsbenchmark,hsieh2024ruler,kim2025rulermeasureallbenchmarking,an_l-eval:_2023}  for LLM acceleration misrepresent real-world dialogue scenarios.
Most benchmarks~\citep{bai2024longbench, li2025exposingnumeracygapsbenchmark,longchat2023} assume queries are always placed at the end of the input and focus on single-turn tasks, which oversimplifies the problem and favors acceleration methods that exploit positional biases. As a result, approaches~\citep{zhang2023h2o, li2024snapkv, feng2024ada} 
that perform well on these benchmarks often fail to generalize to realistic dialogue scenarios where queries may appear at arbitrary positions and contextual dependencies span multiple turns.

In this paper, we propose LoopServe, an adaptive dual-phase LLM inference acceleration framework specifically designed for multi-turn dialogues. LoopServe features two core innovations: online prefilling sparsification and progressive KV compression. In the prefilling phase, LoopServe dynamically identifies and selects the most critical components of the attention matrix, focusing on the vertical and slash line patterns that contribute most to attention weights. Unlike fixed sparsification methods, LoopServe adapts in real time to maintain both efficiency and high attention fidelity. During decoding, LoopServe applies progressive KV compression by dynamically selecting and compressing relevant input tokens based on the most recently generated outputs. This strategy keeps the KV cache efficient and relevant throughout decoding, significantly reducing computational overhead without compromising output quality.
We also introduce a multi-turn long-context benchmark containing 11 datasets. This benchmark captures diverse query positions and multi-turn dependencies, offering a more realistic evaluation framework for dialogue scenarios.
The contributions of this paper are summarized as follows:
%
%
%
%
%
%
\begin{itemize}[leftmargin=*]
	\item We  empirically reveal  that attention patterns and key token positions in multi-turn dialogues are highly dynamic, limiting static sparsification and KV selection.
	
	\item We present LoopServe, a dual-phase LLM acceleration framework with online attention sparsification and progressive KV compression, improving multi-turn inference efficiency.
	
	\item We introduce a benchmark of 11 long-context  multi-turn datasets with varied query positions and dependencies for realistic evaluation.
	
	\item Experiments on    11 multi-turn datasets demonstrate the superior performance of LoopServe.

\end{itemize}

\section{Preliminary and Related Work }\label{sec:related_work}

\subsection{Large Language Models}
\label{ssec:llm}
LLMs like GPT~\citep{NEURIPS2020_1457c0d6}, Llama~\citep{2024llama3herdmodels}, and DeepSeek~\citep{deepseekv3,deepseekr1} excel at context understanding and reasoning, enabled by large-scale training and the Transformer architecture~\citep{vaswani2017attention}. Transformers are effective due to Multi-Head Self-Attention (MHSA), which captures both local and global token dependencies.
Given an input sequence $X = [x_1, x_2, \cdots, x_n]$ with embeddings $\mathbf{X} \in \mathbb{R}^{n \times d}$, the MHSA computes query vectors $\mathbf{Q}^i \in \mathbb{R}^{n \times d_k}$, key vectors $\mathbf{K}^i \in \mathbb{R}^{n \times d_k}$, 
and value vectors $\mathbf{V}^i \in \mathbb{R}^{n \times d_v}$ for the $i$-th attention head as $	\mathbf{Q}^i = \mathbf{X}\mathbf{W}_{Q^i}, 
\mathbf{K}^i = \mathbf{X}\mathbf{W}_{K^i}, 
\mathbf{V}^i = \mathbf{X}\mathbf{W}_{V^i}$,
where $\mathbf{W}_{Q^i}$, $\mathbf{W}_{K^i}$, and $\mathbf{W}_{V^i}$ are learnable matrices. 
Each   $i$-th attention head $\mathbf{Z}^i$ as :
$
	\mathbf{Z}^i = \textsf{Attention}(\mathbf{Q}^i, \mathbf{K}^i, \mathbf{V}^i) = \textsf{Softmax}\left(\frac{\mathbf{Q}^i (\mathbf{K}^i)^\top}{\sqrt{d_k}}\right) \mathbf{V}^i.
$
Next,
outputs from $h$ heads are concatenated and projected:
$\mathbf{Z}=\textsf{Concat}(\mathbf{Z}^1, \mathbf{Z}^2, \dots, \mathbf{Z}^h)\mathbf{W}_O,$
where $\mathbf{W}_O$ is a learned projection. 
For text generation, LLMs use an autoregressive process: given $X = [x_1, \ldots, x_n]$, the model predicts the next token $x_{n+1}$ by modeling
$P(x_{n+1} | x_1, x_2, \cdots, x_n) = \textsf{Softmax}(\mathbf{h}_n \mathbf{W}_{\text{out}} + \mathbf{b}_{\text{out}})$,
where $\mathbf{h}_n$ is the state at step $n$. The next token $x_{n+1}$ is sampled from this distribution and appended to the sequence. Generation continues until an end-of-sequence token or a maximum length is reached.

\subsection{Efficient Long-Context Inference}
\label{ssec:long_context_methods}

The performance of LLMs degrades with long input contexts, due to
the quadratic complexity of self-attention, which scales as $O(Lhn^2)$ for $L$ layers, $h$ heads, and sequence length $n$~\citep{journals/corr/abs-2412-19442}. It makes long-sequence processing prohibitively expensive.

\noindent\textbf{Context Compression Methods.}
Context compression methods reduce the effective sequence length, transforming lengthy inputs into more manageable representations. Filtering-based approaches such as LLMLingua~\citep{jiang2023llmlingua}, LLMLingua-v2~\citep{pan2024llmlingua}, and CompAct~\citep{yoon2024compact} focus on identifying and preserving high-relevance content, allowing models to process only critical information. In contrast, RAG-based (retrieval-augmented generation) methods~\citep{zhao2024chat2data,DBLP:journals/pvldb/JiangZWHA24,wang2025graph,chen2025automatic,edge2024local} construct knowledge graphs or extract semantic triples from the input, synthesizing them into condensed forms for LLMs. These strategies substantially decrease computational and memory costs, but may sacrifice fine-grained details, potentially affecting output quality.
%
%

Also, to reduce computational burden, KV-based approaches minimize the number of attention weight calculations during both prefilling and decoding,  summarized in Table~\ref{tab:long_context_methods} in the Appendix.

\fakeparagraph{Prefilling-stage Optimization}
Self-attention requires $O(n^2)$ computation for an input of length $n$. Recent methods such as Minference~\citep{jiang2024minference}, FlexPrefill~\citep{lai2025flexprefill}, and CritiPrefill~\citep{lv2024critiprefill} use binary  masks $\mathbf{M} \in \{0, 1\}^{n \times n}$ to zero out less important attention weights:
$	 \min \, \left| \mathbf{{A}}^j_i -  \mathbf{\hat{A}}^j_i  \right|,
s.t.\ ,\mathbf{\hat{A}}^j_i = \textsf{Softmax}\left( \frac{ \mathbf{Q}^j(\mathbf{K}^j)^\top }{\sqrt{d_k}}- c(1-\mathbf{M}) \right)$, where $c$ is a large constant that suppresses masked entries. This reduces complexity to $O(\alpha n^2)$ per head, with $\alpha \ll 1$. However,
Minference~\citep{jiang2024minference} and CritiPrefill~\citep{lv2024critiprefill} rely on fixed attention patterns or block selection, while FlexPrefill~\citep{lai2025flexprefill} adjusts sparsity globally with simple heuristics. However, our experiments (Section 3) show that attention patterns are highly input-dependent and dynamic, so these static or coarse methods struggle to adapt in   multi-turn scenarios.


\fakeparagraph{Decoding-stage Optimization}
During autoregressive generation, KV cache methods such as H2O~\citep{zhang2023h2o}, SnapKV~\citep{li2024snapkv}, AdaKV~\citep{feng2024ada}, and others~\citep{DBLP:conf/iclr/Ge0LZ0024,li2024snapkv} select important tokens to store, reducing redundancy. These approaches assume that critical tokens are near the end of the input, performing well on benchmarks like LongBench~\citep{bai2024longbench, bai2024longbench2}, where queries are always placed last. However, as analyzed in Section~\ref{ssec:kp2}, their effectiveness drops when queries appear elsewhere, underscoring the need for adaptive, context-aware KV selection in real dialogue.

\section{Motivational Experiments and Insights}\label{sec:mv}
To clarify the core challenges in accelerating LLM inference for multi-turn dialogues, we conduct motivational experiments in Section~\ref{ssec:kp1} and \ref{ssec:kp2}.   They reveal how attention patterns are dynamically sparse and how query position influences acceleration effectiveness. Building on these findings, we introduce the LoopServe system, specifically designed to address these real-world challenges.


\subsection{Key Point 1: Uncertain Attention Patterns}\label{ssec:kp1}
As investigated previously, attention head matrices are highly sparse~\cite{jiang2024minference}. Existing acceleration methods~\citep{xiao2024efficient, feng2024ada, li2025scbenchkvcachecentricanalysis,li2024snapkv} often rely on sparsifying attention matrices or selecting important KV tokens based on the assumption that attention patterns are fixed and can be identified offline. In reality, these patterns are highly variable across inputs, heads, and layers, limiting the effectiveness of such approaches, as reveled as follows.
 
\begin{figure}[t]
	\centering
	\vspace{-4em}
	\subfloat[\scriptsize{Vertical and slash lines.}]{
		\centering\includegraphics[width=0.18\linewidth]{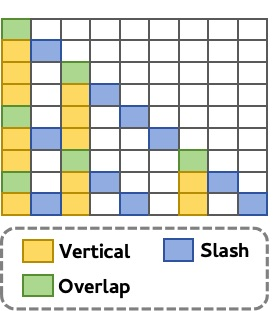}
	}
	\hfill
	\subfloat[\scriptsize{All heads of Llama 3.1.}]
	{\centering\includegraphics[width=0.27\linewidth]{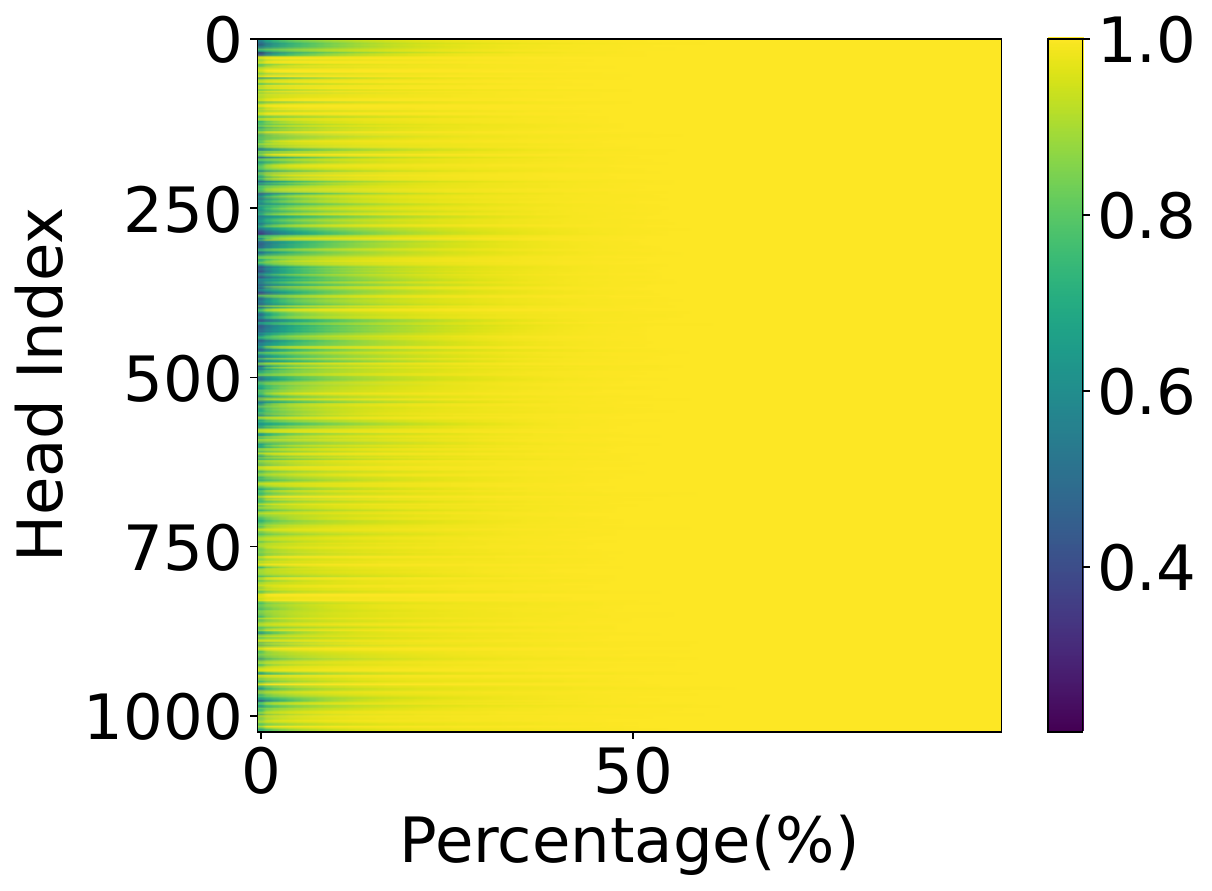}}
	\hfill
	\subfloat[\scriptsize{Average ratio on all heads.}]		
	{\centering\includegraphics[width=0.25\linewidth]{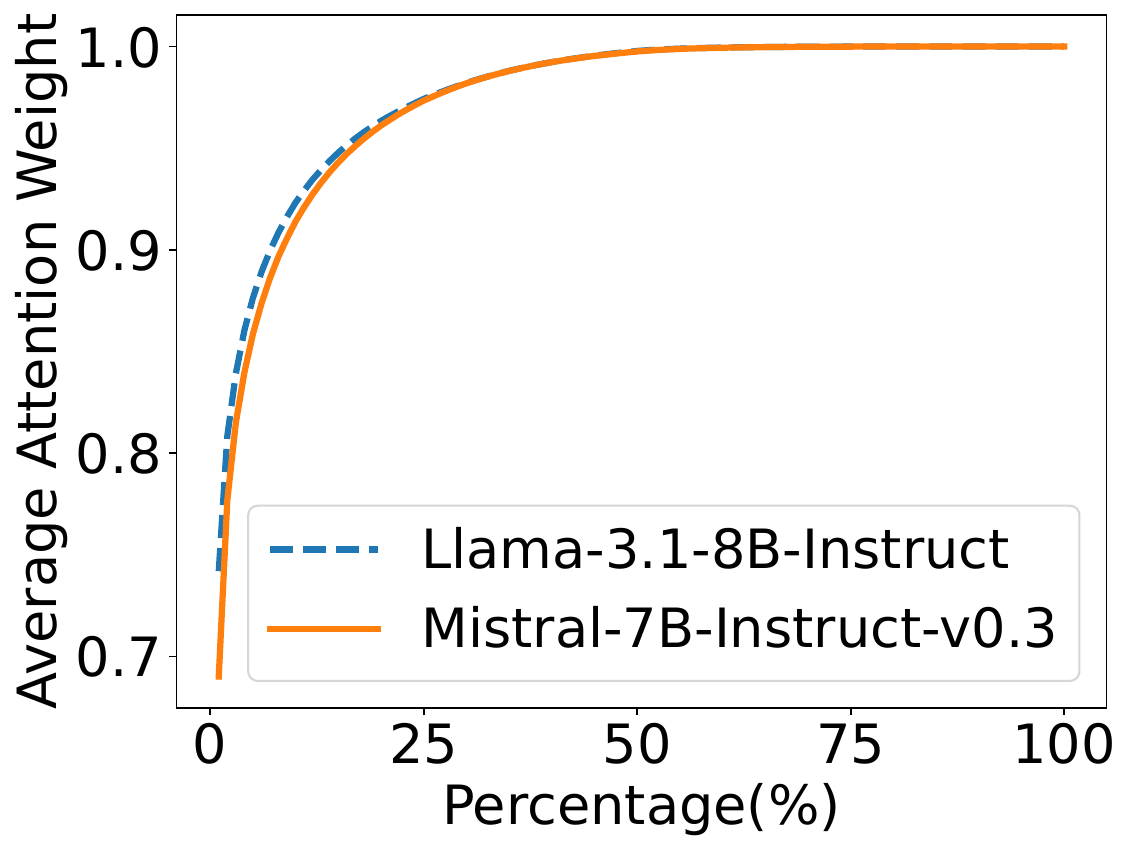}}	
	\hfill
	\subfloat[\scriptsize{Overlap among different inputs.}]{
		\centering\includegraphics[width=0.251\linewidth]{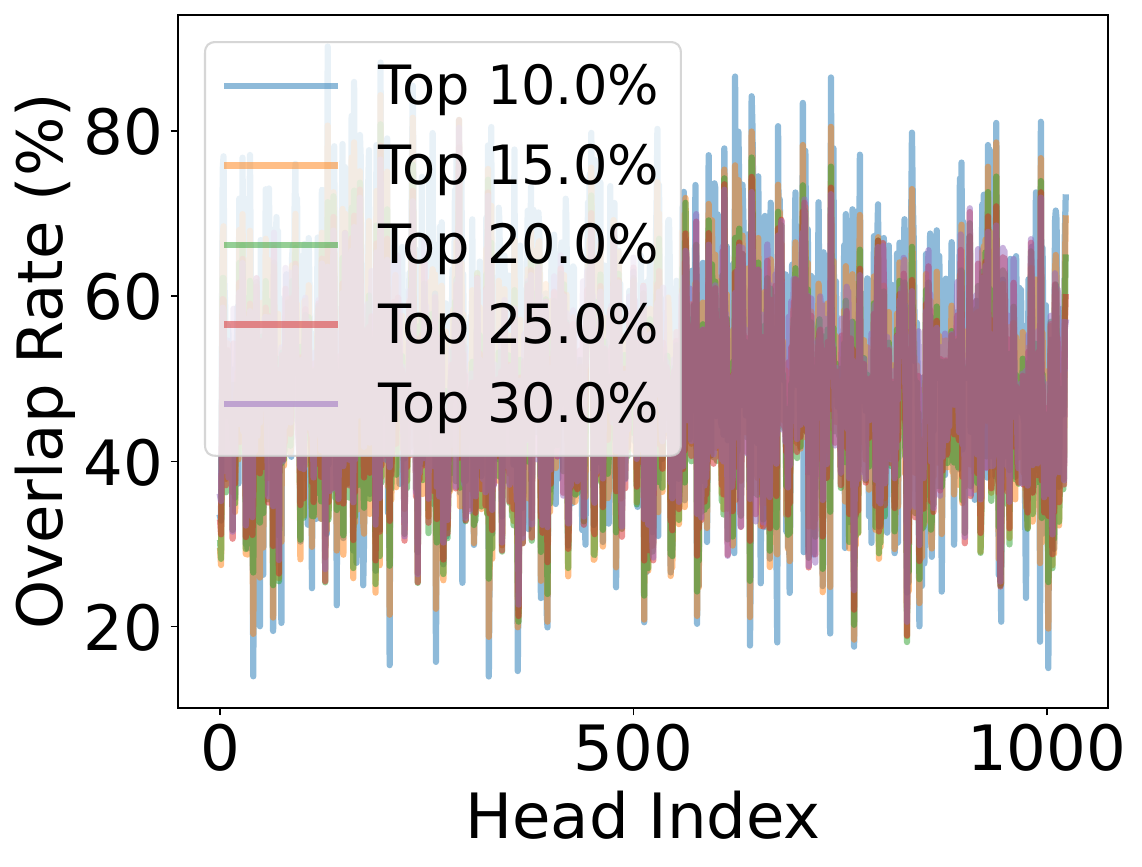}
	}
	\hfill
	\vspace{-1em}
	\caption{Attention head sparsity is shown in (b) and (c). } 
\label{fig:mv1_sparsity_and_mv2_diff}
\vspace{-2em}
\end{figure}

\noindent{\textbf{Motivational Observation 1:	 Only 10\% of vertical and slash lines can collectively account for most (e.g., 90\%) of the attention weight.}
	As shown in Figure~\ref{fig:mv1_sparsity_and_mv2_diff}~(a),
in an attention matrix, a vertical line (column) indicates a single token attended by all others, common for special tokens like separators or keywords. A slash line (diagonal) shows each token mostly focusing on its nearby tokens, reflecting local attention patterns.
We analyze this using the SAMSum QA dataset~\citep{bai2024longbench} and the Llama-3.1-8B-Instruct with $n_h$  heads. For each query $X_i$ of length $n_i$, each $k$-th attention matrix $\mathbf{A}^k_i$ contains $n_i$ vertical lines $\mathcal{V}^k_i$ and $n_i$ slash lines $\mathcal{S}^k_i$. The total attention weight for a slash line $s^k$ is $\sum_{(a,b) \in s^k} \mathbf{A}^k_i[a][b]$, and for a vertical line $v^k$, it is $\sum_{(a,b) \in v^k} \mathbf{A}^k_i[a][b]$.
We select the top $\eta \cdot 2n_i$ slash and vertical lines ($\mathcal{\hat{S}}^k_i$, $\mathcal{\hat{V}}^k_i$) based on their total weights. For each head $k$, we compute the ratio of  weight within these lines:
$
r^k_i = \frac{1}{|\mathcal{D}|}\sum_{X_i \in \mathcal{D}} \frac{\sum_{(a,b) \in \mathcal{\hat{S}}^k_i \cup \mathcal{\hat{V}}^k_i} \mathbf{A}^k_i[a][b]}{n_i}.
$
Averaging over all $n_h$ heads yields the mean ratio. As shown in Figure~\ref{fig:mv1_sparsity_and_mv2_diff}~(b), higher $\eta$ increases cumulative attention weight. Figure~\ref{fig:mv1_sparsity_and_mv2_diff}~(c) shows that for both Llama and Mistral, just 10\% of slash and vertical lines account for 90\% of total attention, indicating highly concentrated attention and enabling efficient selection by focusing on these sparse lines.

\noindent{\textbf{Motivational Observation 2: The positions of the top vertical and slash lines within the same head vary across different user inputs.}}
For a model $M_\theta$ with $n_h$ attention heads and dataset $\mathcal{D} = {X_i}$, we select the top $\eta \cdot 2n_i$ vertical and slash lines ($\mathcal{\hat{V}}^k_i$, $\mathcal{\hat{S}}^k_i$) for each input $X_i$ and attention head $k$. For any pair of queries $X_i$, $X_j$, the overlap of their selected lines under head $k$ is: $
r^k_{i,j} = \frac{|\mathcal{\hat{S}}_i^k \cap \mathcal{\hat{S}}_j^k| + |\mathcal{\hat{V}}_i^k \cap \mathcal{\hat{V}}_j^k|}{|\mathcal{\hat{S}}_i^k \cup \mathcal{\hat{S}}_j^k| + |\mathcal{\hat{V}}_i^k \cup \mathcal{\hat{V}}_j^k|}.
$
Averaging over all heads and input pairs gives the mean overlap ratio:
$\frac{1}{n_h |\mathcal{D}|^2} \sum_{k=1}^{n_h} \sum_{X_i, X_j \in \mathcal{D}} r^k_{i,j}$.
Using the SAMSum QA dataset~\citep{bai2024longbench} and models Llama-3.1-8B-Instruct and Mistral-7B-Instruct-v0.3, with $\eta$ ranging from 0.1 to 0.3, Figure~\ref{fig:mv1_sparsity_and_mv2_diff}~(d) and Figure~\ref{fig:mv2_diff} in Appendix  show that for most heads, the overlap remains below 0.5. This indicates that the most important lines differ significantly depending on the input, even within the same head. As a result, important lines cannot be reliably determined offline for use during online inference.

\noindent{\textbf{Motivational Observation 3: For an input $X_i = [C_i^1, C^2_i]$ split into two segments, the top vertical and slash lines within the same head differ between $C^1_i$ and $C^2_i$. Each segment shows its own local attention sparsity pattern.}}
As illustrated in Figure~\ref{fig:mv2_same_x_and_mv4_question_position}~(a), the key vertical and slash lines in $C^1_i$'s attention matrix are largely absent in $C^2_i$, which displays distinct local patterns. To verify this, we use the SAMSum QA~\citep{bai2024longbench} dataset and Llama-3.1-8B-Instruct.
For each $X_i$ in dataset $\mathcal{D}$, we split it into $[C^1_i, C^2_i]$ and extract the attention matrices $\mathbf{A}^k_{C^1_i}$ and $\mathbf{A}^k_{C^2_i}$ for each head $k$. After selecting the top-$\eta$ important slash and vertical lines for each ($L^k_{C^1_i}$, $L^k_{C^2_i}$) for ( $\mathbf{A}^k_{C^1_i}$, $\mathbf{A}^k_{C^2_i}$), 
we compute the overlap rate:
$r^k_{C_i^1\to C_i^2}= {\sum_{l \in L^k_{C^1}} \mathbb{I}(l \in L^k_{C_i^2})}/{|L^k_{C_i^2}|},$
where $\mathbb{I}$ indicates whether a line from $C^1_i$ is also important in $C^2_i$. 
Averaging across data gives the mean overlap line rate for each head.
Figure~\ref{fig:mv2_same_x_and_mv4_question_position}~(b) shows that, for different $\eta$, the overlap in important lines between $C^1_i$ and $C^2_i$ is consistently low and unstable. This confirms that each segment exhibits unique local attention patterns.
This finding indicates that using only a segment (such as the last window or last few tokens) to predict important attention patterns for the whole input is unreliable. As a result, acceleration methods like Minference~\citep{jiang2024minference}, SnapKV~\citep{li2024snapkv}, H2O~\citep{zhang2023h2o}, and Keyformer~\citep{adnan2024keyformer}, which rely on such assumptions, struggle to deliver consistent performance in real-world scenarios.

\subsection{Key Point 2: Question Position Matters.}\label{ssec:kp2}
\textit{\textbf{Motivational experiments indicate that both prefilling based methods and decoding phase acceleration methods, which depend on offline sparse pattern discovery or fixed sparse patterns, tend to underperform in practical scenarios. However, their reported outcomes on existing benchmarks are often similar to those of large language models that use full attention. Why does this discrepancy occur?}}
The main reason is that benchmarks like Longbench~\citep{bai2024longbench,bai2024longbench2} always place the user question $q_i$ at the end of the input $X_i=[C_i, q_i]$, so the LLM answers $q_i$ based on context $C_i$. In this setup, acceleration methods only need to focus on the last observation window (near the question), which makes it easier to identify context tokens relevant to the question, thus partially mitigating the unpredictability found in real-world input patterns.

%
\begin{figure}[t]
	\centering
	\vspace{-4em}
	\subfloat[\scriptsize{Attention map.}]{
		\centering\includegraphics[width=0.254\linewidth]{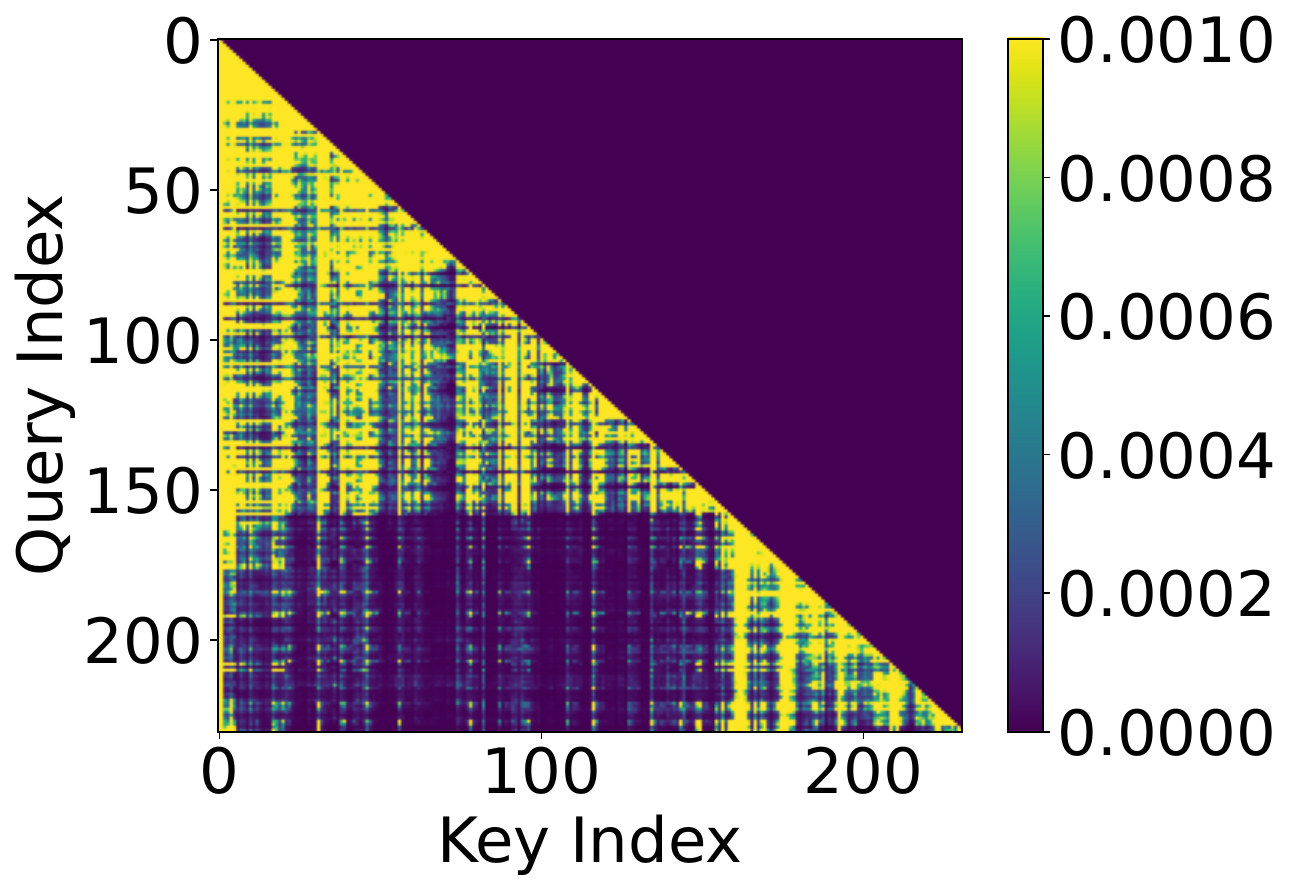}
	}
	\hfill
	\subfloat[\scriptsize{Overlap of different segments.}]{
		\centering\includegraphics[width=0.235\linewidth]{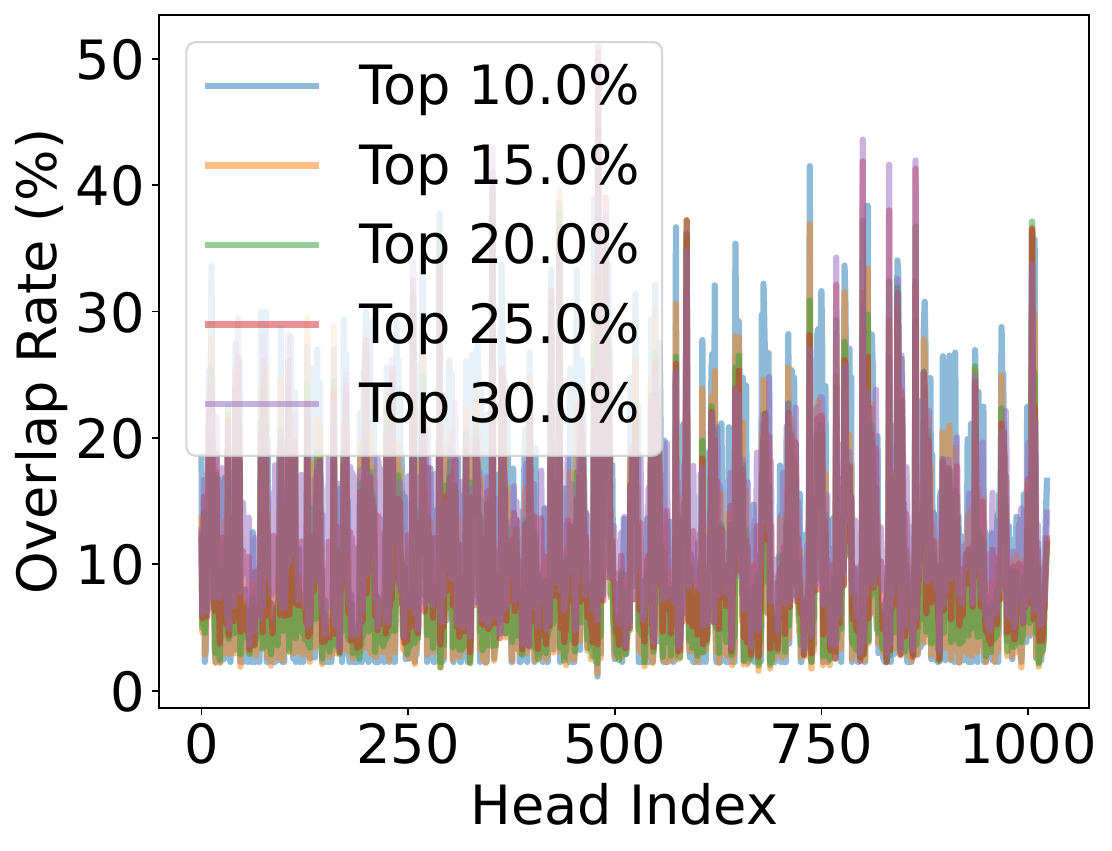}
	}
	\hfill
	\subfloat[\scriptsize{Important token overlap.}]		
	{\centering\includegraphics[width=0.245\linewidth]{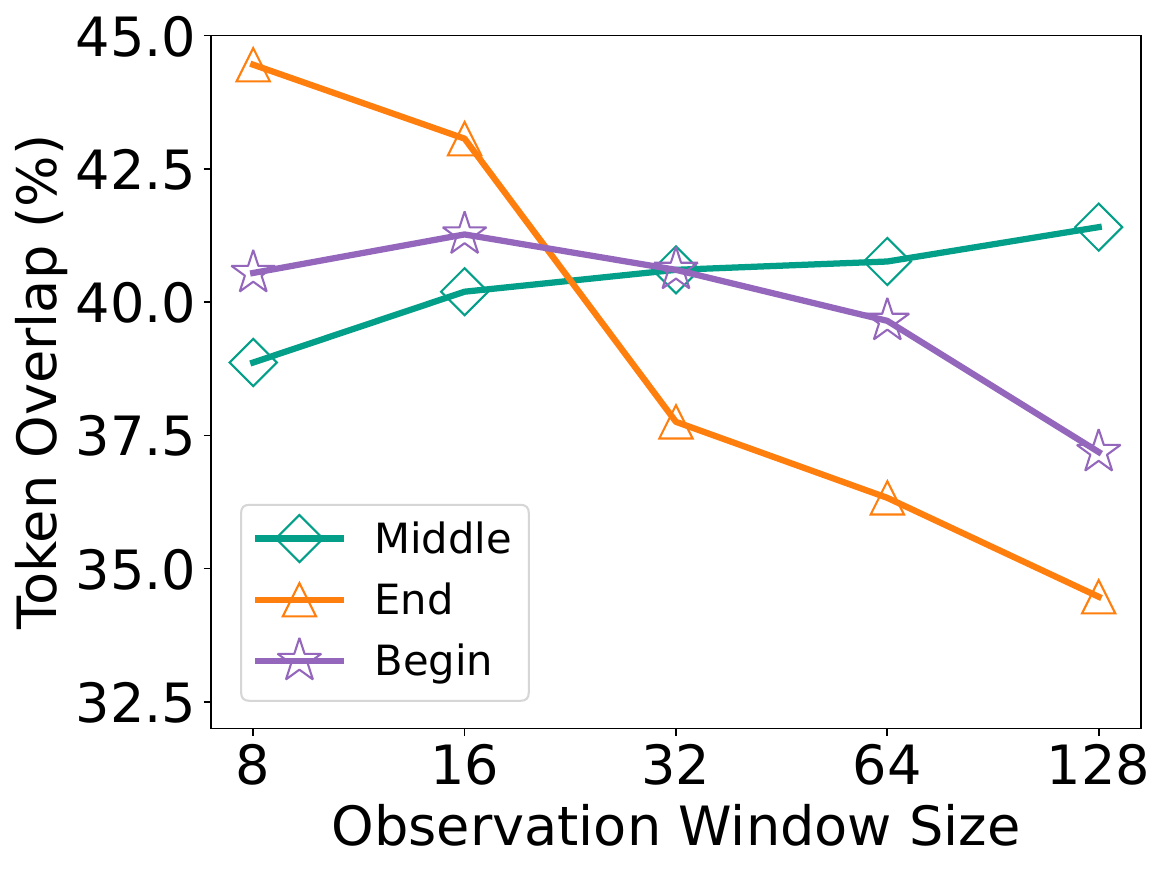}}	
	\hfill
	\subfloat[\scriptsize{Question in different position.}]
	{\centering\includegraphics[width=0.24\linewidth]{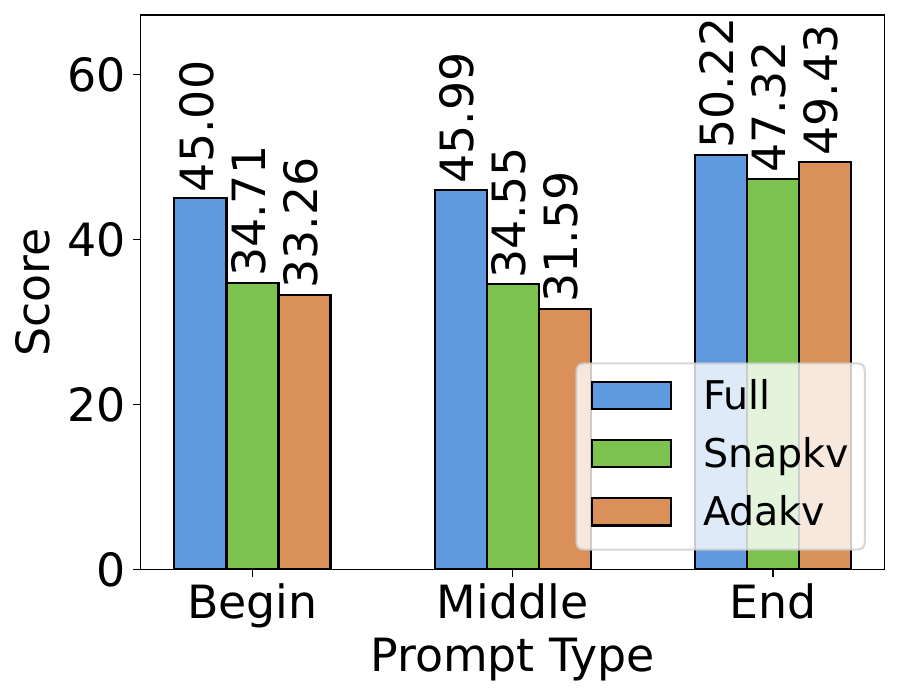}}
	\caption{Different attention sparsity patterns and query position impact on performance.}
\label{fig:mv2_same_x_and_mv4_question_position}
\end{figure}

\noindent{\textbf{Motivational Observation 4: Relying only on the last observation window cannot reliably identify important input tokens for generating the output.}}
Recent methods like H2O~\citep{zhang2023h2o}, SnapKV~\citep{li2024snapkv}, and AdaKV~\citep{feng2024ada} select the top-$B$ important tokens $\hat{X}_i$ from $X_i$ based on attention between each input token and the last observation window $X^{obs}_{i}$ (the last $n_s$ tokens), formally as $	\hat{X}_i = \arg\max_{\hat{X}_i \subseteq X_i} \sum_{j=1}^{n_h} \sum_{a \in \hat{X}_i} \sum_{b \in X^{obs}_i} \mathbf{A}^k_i[a][b]$.
However, the true top-$B$ tokens $\hat{X}^*_i$ for generating the output $Y_i$ should be selected based on their attention to the output tokens: $\hat{X}^*_i = \arg\max_{\hat{X}_i \subseteq X_i} \sum_{j=1}^{n_h} \sum_{a \in \hat{X}_i} \sum_{b \in Y_i} \mathbf{A}^k_i[a][b]$.
We measure the overlap $r_i$ between $\hat{X}_i$ and $\hat{X}^*_i$ ($B$ is set to 10\% of $|X_i|$), using Llama-3.1-8B-Instruct and LongEval’s topic retrieval set, where the instruction $q_i$ is placed at the beginning, middle, or end of $C_i$. 
As shown in Figure~\ref{fig:mv2_same_x_and_mv4_question_position}~(c),   the average overlap of important tokens is highest when the question is at the end, but much lower when it appears in middle or the beginning. This demonstrates that focusing only on the last part of the input misses relevant information unless the question is placed last.

Similarly, as shown in Figure~\ref{fig:mv2_same_x_and_mv4_question_position}~(d), SnapKV and AdaKV match the original model only when the question is at the end. Their performance drops sharply when the question appears earlier, since they rely on the last tokens for context selection. This shows that current methods are overly dependent on input order and do not generalize well when question positions vary.

\vspace{-1em}
\section{Multi-turn Long-Context Benchmarks}\label{sec:benchmark}
Existing benchmarks, 
 such as NumericBench~\citep{li2025exposingnumeracygapsbenchmark}, LongBench~\citep{bai2024longbench, bai2024longbench2}, and LongEval~\citep{longchat2023} 
 focus on single-turn tasks and place user queries only at the context end, 
 which do not reflect the complexity of real-world, multi-turn conversations~\citep{li2025scbenchkvcachecentricanalysis}.
We introduce a benchmark of 11 long-context  multi-turn datasets with varied query positions and dependencies.
Specifically, each $m$-turn instance is defined as
$\mathcal{D} = {\mathcal{I}_i = [(C_{i,1}, q_{i,1}, a_{i,1}), \dots, (C_{i,m}, q_{i,m}, a_{i,m})]}_{i=1}^{|\mathcal{D}|}$,
where $C_{i,j}$ is the context at turn $j$ (possibly empty), $q_{i,j}$ is the user query, and $a_{i,j}$ is the LLM-generated answer.
We ensure diversity by: (1) \textbf{Query Position:} For each turn, the query $q_{i,j}$ can appear at the beginning, end, or between paragraphs of $C_{i,j}$, reflecting more realistic query placements.
(2)  \textbf{Query Relevance:} Answers $a_{i,j}$ may depend on any subset of current and previous contexts $\{C_{i,1},\ldots,C_{i,j}\}$, with variable subset sizes, simulating diverse real-world dependencies.
We construct multi-turn benchmarks for several tasks, including question answering, summarization, and few-shot learning. For construction procedures and detailed benchmark statistics, please refer to Appendix~\ref{appx:ssec:longbench}.

\setlength{\textfloatsep}{3pt}

\section{LoopServe System}\label{sec:method}

\begin{figure}
		\vspace{-2em}
	\centering
	\includegraphics[width=\textwidth]{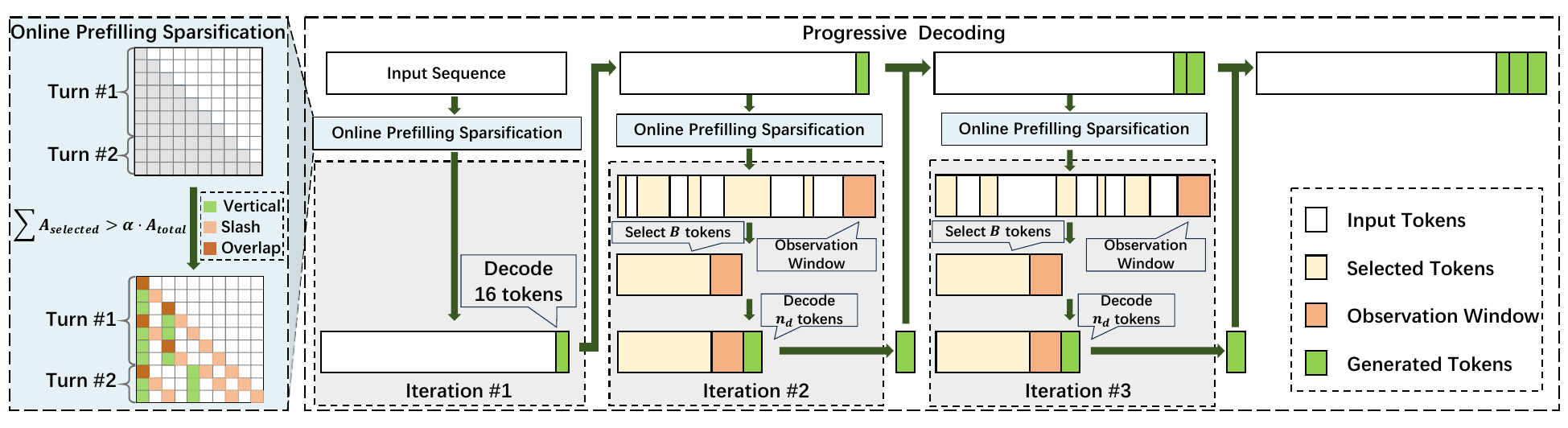}
	\label{fig:loopserve_system}
	\vspace{-2em}

	\caption{Framework overview of LoopServe.}
		\label{fig:framework}
\end{figure}

 As shown in Figure~\ref{fig:framework},
we propose LoopServe, an adaptive dual-phase system that performs online attention sparsification during the prefilling phase and progressive KV compression during the decoding phase. 
%
For an $m$-turn input $\mathcal{I}_i = \{X_{i,j}\}_{j=1}^m$, where $X_{i,j}$ is the context or query in the $j$-th turn, LoopServe generates each answer $y_{i,j}$ using these two steps.

%
%
%
%
%

\noindent \textbf{Step 1. Online Attention Head Sparsification in Prefilling.}
In Algorithm~\ref{alg:framwork} (line 3-5) in Appendix, 
for each new input $X_{i,j}$ and all inputs $X_i=\left(\cup_{j'=1}^{j-1}{(X_{i,j'}\cup y_{i,j'})}\right)\cup X_{i,j}$, 
we get  $\hat{X}_{i,j}=[y_{i,j-1}, X_{i,j}]$ as the new appended input in the $j$ turn.
Then, for each $k$-th attention head, we first compute the attention matrix $\mathbf{A}^k_i[\hat{X}_{i,j}] \in \mathbb{R}^{\hat{n}_{i,j} \times n_i}$, 
and then select the slash lines $\mathcal{\hat{S}}^k_{i,j}$ and vertical lines  $\mathcal{\hat{V}}^k_{i,j}$ that collectively recover at least $\alpha$ of the total attention weight in $\mathbf{A}^k_i[\hat{X}_{i,j}]$.
 
\noindent \textbf{Step 2. Progressive KV Compression in Decoding.}
As described in Algorithm~\ref{alg:framwork} (lines 7–14) in Appendix, 
  after every re-selection interval $n_d$ tokens, the framework uses the \textsf{ProgressiveSelection} Algorithm~\ref{alg:decoding} to compute a subset of input tokens $\hat{X}^k_i \subseteq X_i$ for each attention head $k$.
 these selected tokens $\hat{X}^k_i $ are important for output generation. At each decoding step, LoopServe leverages the compressed KV cache $\{\mathcal{\hat{S}}^{k}_{i,j'}, \mathcal{\hat{V}}^{k}_{i,j'}\}_{j'=1, k}^{j, n_h}$ to generate the output sequence $y_{i,j}$.

\subsection{Online Attention Sparsification in Prefilling}
As shown in \textit{\textbf{Key Point 1}} in Section~\ref{ssec:kp1}, attention sparsity patterns are highly dynamic, making static or offline selection ineffective.
To address this, we propose an online adaptive algorithm that, during prefilling, selects a subset of slash and vertical lines to recover at least an $\alpha$ fraction of the total attention weight for each head, which can be reused     in later dialogue turns.


\begin{definition}[Online Prefilling Sparsification Problem]
	Given the   LLM model $M_\theta$ with $n_h$ attention heads, the input 
	$X_i = \left(\bigcup_{j'=1}^{j-1} (X_{i,j'} \cup y_{i,j'})\right) \cup X_{i,j}$,
	where $y_{i,j'}$ is the answer for $X_{i,j'}$ and $X_{i,j}$ is the current turn’s input. 
	We denote the concatenation of the previous answer $y_{i,j-1}$ and the current user input $X_{i,j}$ as $\hat{X}_{i,j} = [y_{i,j-1}, X_{i,j}]$, whose corresponding attention matrix requires sparsification.
	The $k$-th attention matrix between $X_{i}$ and $\hat{X}_{i,j}$ is denoted as
	$\mathbf{A}^k_{i}[\hat{X}_{i,j}] \in \mathbb{R}^{\hat{n}_{i,j} \times n_i}$.
	Let $\mathcal{S}^k_{i,j}$ (resp.,  $\mathcal{V}^k_{i,j}$ ) denote the set of all  slash lines (resp., vertical lines) in $\mathbf{A}^k_{i}[\hat{X}_{i,j}] $. 
	The goal is to select a subset of slash lines $\hat{\mathcal{S}}^k_{i,j} \subseteq \mathcal{S}^k_{i,j}$ and a subset of vertical lines $\hat{\mathcal{V}}^k_{i,j} \subseteq \mathcal{V}^k_{i,j}$ such that together they recover at least an $\alpha$ fraction of the total attention weight in $\mathbf{A}^k_{i}[\hat{X}_{i,j}]$, where $\alpha \in [0, 1]$.
	\begin{align}\label{eq:prefilling_obj}
		\min\sum_{s \in \mathcal{\hat{S}}^k_{i,j}} l_s + \sum_{v \in \mathcal{\hat{V}}^k_{i,j}} l_v,
		\quad	s.t.   \sum_{(a, b) \in (\mathcal{\hat{S}}^k_{i,j} \cup \mathcal{\hat{V}}^k_{i,j})} \mathbf{A}^k_i[a][b] \geq \alpha \cdot \hat{n}_{i,j},
	\end{align}
	where $\hat{n}_{i,j}$  is the total attention weight of the matrix $\mathbf{A}^k_{i}[\hat{X}_{i,j}]$, and $l_s$ (resp., $l_v$) is the length of the slash line $s$ (resp., vertical line $v$).
\end{definition}

\begin{theorem}
	The  prefilling sparsification problem is NP-hard. The proof  is detailed in Appendix~\ref{ssec: proof of theorem 1}. 
\end{theorem}

\noindent\textit{\textbf{Algorithm.}}
Algorithm~\ref{alg:prefilling} in Appendix~\ref{ssec:alg} takes as input the concatenated sequence $\hat{X}{i,j} = [y_{i,j-1}, X_{i,j}]$, where $y_{i,j-1}$ is the previous answer and $X_{i,j}$ is the current user input. It also requires the $k$-th attention head of the LLM $M_\theta$ and a sparsity threshold parameter $\alpha$. The output consists of the selected slash lines $\mathcal{\hat{S}}^k_{i,j}$ and selected vertical lines $\mathcal{\hat{V}}^k_{i,j}$.
The algorithm begins by sampling a subset $\tilde{X}_{i,j}$ from the concatenated input $\hat{X}_{i,j}$ to reduce computational cost. It then computes the query matrix $\mathbf{\tilde{Q}}^k_{i,j}$ for $\tilde{X}_{i,j}$ and the key matrix $\mathbf{K}^k_i$ for the full input $X_i$.
Next, all slash lines $\mathcal{S}^k{i,j}$ and vertical lines $\mathcal{V}^k_{i,j}$ are summarized based on $\mathbf{A}^k_i[\tilde{X}{i,j}]$ and sorted in descending order. Two empty sets, $\mathcal{\hat{S}}^k_{i,j}$ and $\mathcal{\hat{V}}^k_{i,j}$, are initialized to store the selected lines, while the overlap weights $ol_s$ and $ol_v$ are initialized to zero.
Algorithm~\ref{alg:prefilling} then iteratively selects lines until the total recovered attention weight $\texttt{sum}$ meets or exceeds $\alpha \cdot \textsf{sum}(\mathbf{A}^k_i[\tilde{X}_{i,j}])$. At each iteration, it compares the top slash line $s \in \mathcal{S}^k_{i,j}$ and the top vertical line $v \in \mathcal{V}^k_{i,j}$ based on their marginal contributions, $\Delta w_s = w_s - ol_v$ and $\Delta w_v = w_v - ol_s$. Since each slash line overlaps with only one vertical line, $\Delta w_s = w_s - ol_v$.
The line with the greater marginal contribution is added to its respective set ($\mathcal{\hat{S}}^k_{i,j}$ or $\mathcal{\hat{V}}^k_{i,j}$), and the overlap weights and total recovered weight are updated accordingly.
The loop terminates once the recovery condition is satisfied, and the algorithm returns the sets $\mathcal{\hat{S}}^k_{i,j}$ and $\mathcal{\hat{V}}^k_{i,j}$. The time complexity is $O(n_i|\tilde{X}_{i,j}| + n_i \log n_i + n_i)$, as detailed in Appendix~\ref{sssec:time_complexity}.

\subsection{Progressive KV Compression in Decoding}

As shown in Section~\ref{ssec:kp2}, fixed or last-token-focused decoding struggles when queries are not at the end of the input. To overcome this, we propose a progressive KV compression that selects important input tokens based on recent outputs, leading to greater overlap with truly important input tokens.


To verify this, we consider the LLM $M_\theta$ with $n_h$ attention heads, an input sequence $X_i$, a generated output sequence $Y_i$, and a window size $n_w$.
As in Section~\ref{ssec:kp2}, we compute the ground-truth top-$B$ important tokens $\hat{X}^*_i \subseteq X_i$ using the output $Y_i$. 
We then use observation windows, extracted from either $X_i$ or $Y_i$, to select the top-$B$ important tokens.
Specifically, given the window size $n_w$, the $-\frac{|X_i|}{n_w}$-th to $-1$ observation windows from $X_i$ are defined as
$X_i[n_i-m \cdot n_w:n_i-(m-1) \cdot n_w]$, $X_i[n_i-(m-1) \cdot n_w:n_i-(m-2) \cdot n_w]$, ..., $X_i[n_i-n_w:n_i]$.
The $0$-th to $\frac{|Y_i|}{n_w}$ observation windows from $Y_i$ are
$Y_i[0 : n_w]$, $Y_i[n_w : 2n_w]$, ..., $Y_i[(m-1) \cdot n_w : m \cdot n_w]$.
For each observation window $X^{obs}_i$ from $X_i$ or $Y_i$, we select the top-$B$ tokens $\hat{X}_i \subseteq X_i$ following Section~\ref{ssec:kp2}, and compute the overlap rate between $\hat{X}_i$ and the ground truth $\hat{X}^*_i$ as $\frac{|\hat{X}_i \cap \hat{X}^*_i|}{B}$.

\begin{wrapfigure}{r}{0.5\textwidth}
	\centering
	\subfloat[\scriptsize{Important token overlapping.}]		
	{\centering\includegraphics[width=0.47\linewidth]{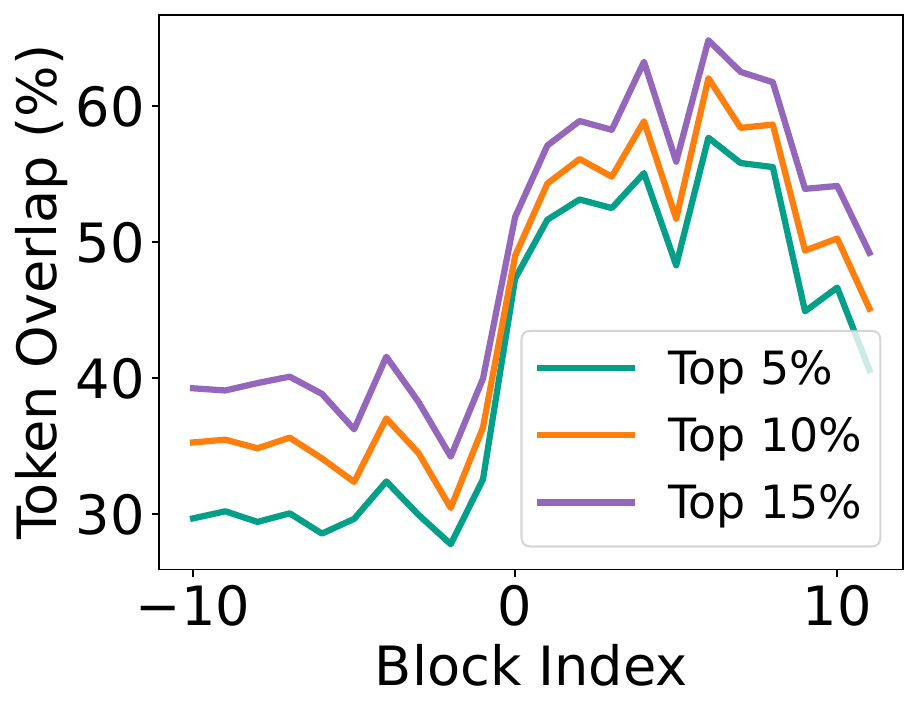}}	
	\hfill
	\subfloat[\scriptsize{Decoding blocks.}]		
	{\centering\includegraphics[width=0.49\linewidth]{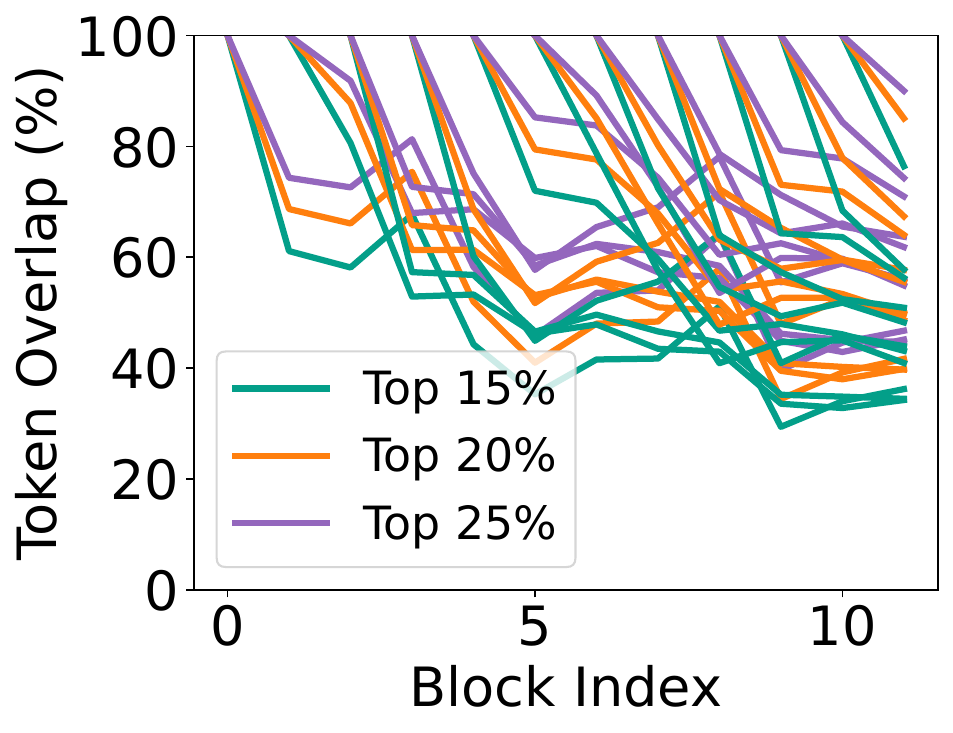}}
	\vspace{-8pt}
	\caption{Progressive decoding. 
	}
	\label{fig:mv5_decoding}
\end{wrapfigure}
We use LongEval~\citep{longchat2023} as in Section~\ref{sec:mv}, experimenting with Llama-3.1-8B-Instruct and setting $B \in \{5\%, 10\%, 15\%\} \cdot |X_i|$. Figure~\ref{fig:mv5_decoding}(a) shows that important tokens selected from input blocks (block index $< 0$) have low overlap rates, while those selected from output blocks ($\geq 0$) have much higher overlap. This indicates that using output tokens is more effective for identifying relevant input tokens.

We also observe that the overlap of important input tokens between output blocks decreases as the distance between blocks increases, which indicates that we can use the most recent output tokens to select important input tokens for the next output block.
Specifically, given the output sequence $Y_i$, we divide it into $n_b$ blocks, i.e., $Y_i = [Y^1_i, \cdots, Y^{n_b}_i]$, where $Y^j_i$ denotes the $j$-th output block of size $\frac{|Y_i|}{n_b}$. For each block $Y^j_i$, we compute the top-$B$ important input tokens as:
$	\hat{X}_{i,Y^j_i} = \arg\max_{\hat{X}_i \subseteq X_i} \sum_{k=1}^{n_h} \sum_{a \in \hat{X}_i} \sum_{b \in Y^j_i} \mathbf{A}^k_i[a][b]$.
Next, we compare the overlap $\frac{|\hat{X}_{i,Y^j_i} \cap \hat{X}_{i,Y^{j'}_i}|}{B}$ of important input tokens between every pair of output blocks $Y^j_i$ and $Y^{j'}_i$. As shown in Figure~\ref{fig:mv5_decoding}~(b), for each block $Y^j_i$, the overlap of important input tokens between $Y^j_i$ and $Y^{j'}_i$ (where $j' > j$) gradually decreases as the block distance increases. Notably, the overlap between $Y^j_i$ and its immediate successor $Y^{j+1}_i$ is higher compared to earlier blocks such as $Y^{j-2}_i$. This indicates that tokens identified from $Y^j_i$ are highly relevant for the generation of $Y^{j+1}_i$.
Therefore, when generating  $Y^{j+1}_i$, we can use the preceding block $Y^j_i$ to dynamically identify the top-$B$ important input tokens.
Based on this, we propose the following progressive KV compression algorithm.

 \fakeparagraph{Algorithm}
 Algorithm~\ref{alg:decoding} firstly set the answer $y_{i,j}$ as empty and decoding step counter $n_o$ to 0 (line 1). 
 During the decoding loop, if the current step reaches either the empirically predefined decoding size  16 or a re-selection interval $n_d$ (line 3), 
 the algorithm extracts the most recent tokens from the input sequence $X_i$, i.e., $X^{obs}_i = X_i[|X_i|-n_d: |X_i|]$), and dynamically updates the compressed KV cache subset $\hat{X}^k_i$ for each $k$-th attention head (line 4-5). Specifically, for each head, we select top-$B$ tokens for each head as:
 $
 	\hat{X}^k_i  = \arg\max_{\hat{X}^k_i \subseteq X_i, |\hat{X}^k_i| = B|} \sum_{a \in \hat{X}_i} \sum_{b \in X^{obs}_i} \mathbf{A}^k_i[a][b],
$
 The LLM generates the next token using the compressed KV cache and the updated input, which is appended to both the input and output sequences (line 6-9). This process iterates until the LLM completes generation, returning the final answer $y_{i,j}$.

\section{EXPERIMENTS}

\subsection{Experimental Settings}\label{ssec:exp:setting}
 
 \textbf{Datasets, Tasks, and Evaluation Metrics.} 
  We design multi-turn long-context benchmarks.
 Each instance contains multiple rounds with diverse query positions and dependencies.
  It covers Question Answering, Summarization, and Few-shot Learning. Dataset statistics and each corresponding metric (e.g., F1, Accuracy, and Rouge-L) are in Table~\ref{multi-turn-dataset-table} in Appendix~\ref{ssec:dataset-generation}.
 

 \textbf{Baselines.} 
 We compare our LoopServe with six state-of-the-art KV cache algorithms on two representation LLM base models,
 including Llama-3.1-8B-Instruct~\citep{2024llama3herdmodels} and Qwen2.5-7B-Instruct~\citep{qwen2.5}. 
 The KV cache methods include SnapKV (Snap)~\cite{li2024snapkv}, AdaKV (Ada)~\cite{feng2024ada}, StreamingLLM (SLLM)~\cite{xiao2024efficient}, A-shape (A-S)~\cite{xiao2024efficient}, Tri-Shape (T-S)~\cite{li2025scbenchkvcachecentricanalysis}, and Minference (Minf)~\cite{jiang2024minference}.
 
 
 \textbf{Hyperparameter and Hardware Setting.} All codes are executed on a Rocky Linux 8.10 machine with an 8-core Intel® Xeon® Gold 6542Y CPU, an NVIDIA H100 GPU with 80GB of memory, and 256GB of RAM. 
 For baselines, we use their suggested setting.
{For main experiments in Section~\ref{ssec:main_results}, for our LoopServe, we set  $\alpha=0.955$, and $n_d=16$ as defaults, and we set the  token budget $B=1024$ following~\cite{li2024snapkv} ~\cite{feng2024ada} for all baselines and LoopServe. 
}

\begin{table*}[t]
	\scriptsize
	\centering
	\vspace{-2em}
	\setlength\tabcolsep{2.5pt}
	\renewcommand{\arraystretch}{1.05}
	
	\caption{Effectiveness Evaluation. The bold number indicates the best performance.}
		\vspace{-1em}
	\label{tab:effectiveness}
	\begin{tabular}{c|cc|cccccccc|cccccccc}
		\hline
		\multirow{2}{*}{\textbf{P}}  & \multicolumn{2}{c|}{\multirow{2}{*}{\textbf{Data}}}              & \multicolumn{8}{c|}{\textbf{Llama-3.1}}                                                                                                                                                                                                                                       & \multicolumn{8}{c}{\textbf{Qwen2.5}}                                                                                                                                                                                                                                            \\ \cline{4-19} 
		& \multicolumn{2}{c|}{}                                            & \multicolumn{1}{c|}{\textbf{Base}} & \multicolumn{1}{c}{\textbf{Snap}} & \multicolumn{1}{c}{\textbf{Ada}} & \multicolumn{1}{c}{\textbf{SLLM}} & \multicolumn{1}{c}{\textbf{A-S}} & \multicolumn{1}{c}{\textbf{T-S}} & \multicolumn{1}{c|}{\textbf{Minf}} & \textbf{Ours} &  \multicolumn{1}{c|}{\textbf{Base}} & \multicolumn{1}{c}{\textbf{Snap}} & \multicolumn{1}{c}{\textbf{Ada}} & \multicolumn{1}{c}{\textbf{SLLM}} & \multicolumn{1}{c} {\textbf{A-S}} & \multicolumn{1}{c}{\textbf{T-S}} & \multicolumn{1}{c|}{\textbf{Minf}} & \textbf{Ours} \\ \hline

		\multirow{11}{*}{\rotatebox{90}{\textbf{Begin}}} & \multicolumn{1}{c|}{\multirow{6}{*}{\rotatebox{90}{\textbf{QA}}}} & \textbf{MFQA-en} & \multicolumn{1}{c|}{45.70} & \multicolumn{1}{c}{36.50} & \multicolumn{1}{c}{34.94} & \multicolumn{1}{c}{25.83} & \multicolumn{1}{c}{28.00} & \multicolumn{1}{c}{34.96} &\multicolumn{1}{c|}{43.05} & \multicolumn{1}{c|}{\textbf{46.82}} & \multicolumn{1}{c|}{44.24} & \multicolumn{1}{c}{35.59} & \multicolumn{1}{c}{33.00} & \multicolumn{1}{c}{21.45} &  \multicolumn{1}{c}{29.56} & \multicolumn{1}{c}{38.14} &\multicolumn{1}{c|}{42.63} & \multicolumn{1}{c}{\textbf{43.47}}         \\
		& \multicolumn{1}{c|}{}                            & \textbf{2WikiMQA} & \multicolumn{1}{c|}{31.68} & \multicolumn{1}{c}{29.31} & \multicolumn{1}{c}{28.73} & \multicolumn{1}{c}{22.05}  & \multicolumn{1}{c}{19.42} & \multicolumn{1}{c}{24.20}& \multicolumn{1}{c|}{28.57} & \multicolumn{1}{c|}{\textbf{35.11}} & \multicolumn{1}{c|}{37.86} & \multicolumn{1}{c}{34.54} & \multicolumn{1}{c}{33.83} & \multicolumn{1}{c}{23.19} & \multicolumn{1}{c}{22.11} & \multicolumn{1}{c}{32.90} & \multicolumn{1}{c|}{35.33} & \multicolumn{1}{c}{\textbf{37.52}}       \\
		& \multicolumn{1}{c|}{}                            & \textbf{Musique} & \multicolumn{1}{c|}{17.16} & \multicolumn{1}{c}{14.29} & \multicolumn{1}{c}{14.80} & \multicolumn{1}{c}{9.76}  & \multicolumn{1}{c}{4.22} & \multicolumn{1}{c}{8.18}& \multicolumn{1}{c|}{14.92} & \multicolumn{1}{c|}{\textbf{18.81}} & \multicolumn{1}{c|}{18.22} & \multicolumn{1}{c}{14.74} & \multicolumn{1}{c}{13.95} & \multicolumn{1}{c}{8.68} & \multicolumn{1}{c}{5.92} & \multicolumn{1}{c}{12.15}  & \multicolumn{1}{c|}{15.15}& \multicolumn{1}{c}{\textbf{16.63}} \\ 
		& \multicolumn{1}{c|}{}                            & \textbf{HotpotQA}& \multicolumn{1}{c|}{36.67} & \multicolumn{1}{c}{33.87} & \multicolumn{1}{c}{32.98} & \multicolumn{1}{c}{22.52} & \multicolumn{1}{c}{12.72} & \multicolumn{1}{c}{21.29} & \multicolumn{1}{c|}{34.38} & \multicolumn{1}{c|}{\textbf{39.00}} & \multicolumn{1}{c|}{42.93} & \multicolumn{1}{c}{38.90} & \multicolumn{1}{c}{37.56} & \multicolumn{1}{c}{21.42} & \multicolumn{1}{c}{17.22} & \multicolumn{1}{c}{31.03} & \multicolumn{1}{c|}{38.67} & \multicolumn{1}{c}{\textbf{42.50}} \\
		& \multicolumn{1}{c|}{}                            & \textbf{NrtvQA}& \multicolumn{1}{c|}{13.40} & \multicolumn{1}{c}{13.42} & \multicolumn{1}{c}{8.67} & \multicolumn{1}{c}{5.22} &  \multicolumn{1}{c}{7.61} & \multicolumn{1}{c}{9.19} &\multicolumn{1}{c|}{13.47} & \multicolumn{1}{c|}{\textbf{14.03}} & \multicolumn{1}{c|}{13.63} & \multicolumn{1}{c}{11.05} & \multicolumn{1}{c}{10.18} & \multicolumn{1}{c}{6.82} & \multicolumn{1}{c}{7.57} & \multicolumn{1}{c}{10.45} & \multicolumn{1}{c|}{10.78} & \multicolumn{1}{c}{\textbf{14.36}} \\
		& \multicolumn{1}{c|}{}                            & \textbf{Qasper} & \multicolumn{1}{c|}{25.33} & \multicolumn{1}{c}{20.77} & \multicolumn{1}{c}{17.03} & \multicolumn{1}{c}{16.58} &\multicolumn{1}{c}{20.64} & \multicolumn{1}{c}{23.37} &  \multicolumn{1}{c|}{22.79} & \multicolumn{1}{c|}{\textbf{24.64}} & \multicolumn{1}{c|}{26.08} & \multicolumn{1}{c}{21.79} & \multicolumn{1}{c}{20.13} & \multicolumn{1}{c}{17.96} & \multicolumn{1}{c}{19.67} & \multicolumn{1}{c}{23.44} &  \multicolumn{1}{c|}{24.92} &\multicolumn{1}{c}{\textbf{25.37}} \\\cline{2-19}

		& \multicolumn{1}{c|}{\multirow{3}{*}{\rotatebox{90}{\textbf{SUM}}}} & \textbf{MultiNews} & \multicolumn{1}{c|}{20.08} & \multicolumn{1}{c}{19.20} & \multicolumn{1}{c}{18.03} & \multicolumn{1}{c}{19.30}  & \multicolumn{1}{c}{20.10} & \multicolumn{1}{c}{20.15}& \multicolumn{1}{c|}{\textbf{20.25}} & \multicolumn{1}{c|}{20.14} & \multicolumn{1}{c|}{18.39} & \multicolumn{1}{c}{15.87} & \multicolumn{1}{c}{14.52} & \multicolumn{1}{c}{17.29}  & \multicolumn{1}{c}{18.29} & \multicolumn{1}{c}{18.30}& \multicolumn{1}{c|}{18.30} & \multicolumn{1}{c}{\color{black}{\textbf{20.14}}} \\
		& \multicolumn{1}{c|}{}                            & \textbf{GovReport} & \multicolumn{1}{c|}{25.25} & \multicolumn{1}{c}{18.36} & \multicolumn{1}{c}{16.53} & \multicolumn{1}{c}{17.98} &\multicolumn{1}{c}{24.16} & \multicolumn{1}{c}{24.13} & \multicolumn{1}{c|}{24.55} &  \multicolumn{1}{c|}{\color{black}{\textbf{24.90}}} & \multicolumn{1}{c|}{20.93} & \multicolumn{1}{c}{16.80} & \multicolumn{1}{c}{15.86} & \multicolumn{1}{c}{15.90} & \multicolumn{1}{c}{22.28} & \multicolumn{1}{c}{21.90} & \multicolumn{1}{c|}{20.86} & \multicolumn{1}{c}{\color{black}{\textbf{23.60}}} \\
		& \multicolumn{1}{c|}{}                            & \textbf{QMSum}& \multicolumn{1}{c|}{20.53} & \multicolumn{1}{c}{17.49} & \multicolumn{1}{c}{17.56} & \multicolumn{1}{c}{14.66} & \multicolumn{1}{c}{17.77} & \multicolumn{1}{c}{18.92} & \multicolumn{1}{c|}{20.15} & \multicolumn{1}{c|}{\textbf{20.65}} & \multicolumn{1}{c|}{19.97} & \multicolumn{1}{c}{17.54} & \multicolumn{1}{c}{17.39} & \multicolumn{1}{c}{14.66} & \multicolumn{1}{c}{18.45} & \multicolumn{1}{c}{19.12} & \multicolumn{1}{c|}{19.61} & \multicolumn{1}{c}{\textbf{19.66}} \\ \cline{2-19}

		& \multicolumn{1}{c|}{\multirow{2}{*}{\textbf{\rotatebox{90}{FS}}}} & \textbf{TREC}& \multicolumn{1}{c|}{46.99} & \multicolumn{1}{c}{46.74} & \multicolumn{1}{c}{45.23} & \multicolumn{1}{c}{46.23} & \multicolumn{1}{c}{41.21} & \multicolumn{1}{c}{46.99} & \multicolumn{1}{c|}{45.48} & \multicolumn{1}{c|}{\textbf{47.99}} & \multicolumn{1}{c|}{65.08} & \multicolumn{1}{c}{64.07} & \multicolumn{1}{c}{63.32} & \multicolumn{1}{c}{62.65} & \multicolumn{1}{c}{60.30} & \multicolumn{1}{c}{56.28} & \multicolumn{1}{c|}{64.07} & \multicolumn{1}{c}{\textbf{65.33}} \\
		& \multicolumn{1}{c|}{}                            & \textbf{SAMSUM} & \multicolumn{1}{c|}{17.32} & \multicolumn{1}{c}{16.75} & \multicolumn{1}{c}{16.48} & \multicolumn{1}{c}{12.95}  & \multicolumn{1}{c}{17.39} & \multicolumn{1}{c}{17.35} & \multicolumn{1}{c|}{17.46}& \multicolumn{1}{c|}{\color{black}{\textbf{18.12}}} & \multicolumn{1}{c|}{16.19} & \multicolumn{1}{c}{13.97} & \multicolumn{1}{c}{13.15} & \multicolumn{1}{c}{11.22}  & \multicolumn{1}{c}{15.32} & \multicolumn{1}{c}{15.92} & \multicolumn{1}{c|}{15.96}& \multicolumn{1}{c}{\color{black}{\textbf{17.15}}} \\ \hline\hline

		\multirow{11}{*}{\rotatebox{90}{\textbf{Middle}}} & \multicolumn{1}{c|}{\multirow{6}{*}{\textbf{\rotatebox{90}{QA}}}} & \textbf{MFQA-en} & \multicolumn{1}{c|}{46.73} & \multicolumn{1}{c}{37.30} & \multicolumn{1}{c}{35.19} & \multicolumn{1}{c}{26.87}  & \multicolumn{1}{c}{28.26} & \multicolumn{1}{c}{33.39} & \multicolumn{1}{c|}{43.84}&  \multicolumn{1}{c|}{\textbf{44.73}}& \multicolumn{1}{c|}{44.00} & \multicolumn{1}{c}{34.21} & \multicolumn{1}{c}{31.68} & \multicolumn{1}{c}{22.73} &  \multicolumn{1}{c}{29.06} & \multicolumn{1}{c}{34.31} &\multicolumn{1}{c|}{41.42} & \multicolumn{1}{c}{\textbf{41.95}} \\
		& \multicolumn{1}{c|}{}                            & \textbf{2WikiMQA} & \multicolumn{1}{c|}{34.10} & \multicolumn{1}{c}{31.71} & \multicolumn{1}{c}{29.90} & \multicolumn{1}{c}{25.59} &  \multicolumn{1}{c}{18.77} & \multicolumn{1}{c}{27.26} &\multicolumn{1}{c|}{32.50} & \multicolumn{1}{c|}{\textbf{34.25}} & \multicolumn{1}{c|}{27.80} & \multicolumn{1}{c}{22.32} & \multicolumn{1}{c}{22.69} & \multicolumn{1}{c}{14.19} & \multicolumn{1}{c}{20.18} & \multicolumn{1}{c}{26.15} &  \multicolumn{1}{c|}{25.40} &\multicolumn{1}{c}{\color{black}{\textbf{27.01}}} \\
		& \multicolumn{1}{c|}{}                            & \textbf{Musique} & \multicolumn{1}{c|}{16.30} & \multicolumn{1}{c}{13.68} & \multicolumn{1}{c}{14.05} & \multicolumn{1}{c}{8.59} &  \multicolumn{1}{c}{3.17} & \multicolumn{1}{c}{9.51} &\multicolumn{1}{c|}{15.39} & \multicolumn{1}{c|}{\textbf{17.48}} & \multicolumn{1}{c|}{10.05} & \multicolumn{1}{c}{7.37} & \multicolumn{1}{c}{7.15} & \multicolumn{1}{c}{3.32} &  \multicolumn{1}{c}{4.92} & \multicolumn{1}{c}{8.38} &\multicolumn{1}{c|}{8.76} & \multicolumn{1}{c}{\textbf{9.13}} \\
		& \multicolumn{1}{c|}{}                            & \textbf{HotpotQA} & \multicolumn{1}{c|}{40.63} & \multicolumn{1}{c}{37.48} & \multicolumn{1}{c}{36.30} & \multicolumn{1}{c}{27.30} &  \multicolumn{1}{c}{12.36} & \multicolumn{1}{c}{25.25} &\multicolumn{1}{c|}{36.81} & \multicolumn{1}{c|}{\textbf{41.25}} & \multicolumn{1}{c|}{29.43} & \multicolumn{1}{c}{25.24} & \multicolumn{1}{c}{24.56} & \multicolumn{1}{c}{12.25} & \multicolumn{1}{c}{12.96} & \multicolumn{1}{c}{22.00} &  \multicolumn{1}{c|}{26.43} &\multicolumn{1}{c}{\textbf{29.05}} \\
		& \multicolumn{1}{c|}{}                            & \textbf{NrtvQA} & \multicolumn{1}{c|}{15.29} & \multicolumn{1}{c}{13.06} & \multicolumn{1}{c}{10.64} & \multicolumn{1}{c}{7.26} & \multicolumn{1}{c}{7.14} & \multicolumn{1}{c}{10.10} & \multicolumn{1}{c|}{13.98} & \multicolumn{1}{c|}{\textbf{15.25}} & \multicolumn{1}{c|}{14.82} & \multicolumn{1}{c}{11.88} & \multicolumn{1}{c}{11.83} & \multicolumn{1}{c}{9.04} & \multicolumn{1}{c}{7.68} & \multicolumn{1}{c}{11.24} & \multicolumn{1}{c|}{11.75} & \multicolumn{1}{c}{\textbf{15.21}} \\ 
		& \multicolumn{1}{c|}{}                            & \textbf{Qasper} & \multicolumn{1}{c|}{30.79} & \multicolumn{1}{c}{26.80} & \multicolumn{1}{c}{22.07} & \multicolumn{1}{c}{21.40} & \multicolumn{1}{c}{24.90} & \multicolumn{1}{c}{28.86} & \multicolumn{1}{c|}{28.47} & \multicolumn{1}{c|}{\textbf{31.12}} & \multicolumn{1}{c|}{28.74} & \multicolumn{1}{c}{23.57} & \multicolumn{1}{c}{21.77} & \multicolumn{1}{c}{20.75} & \multicolumn{1}{c}{24.73} & \multicolumn{1}{c}{27.84} & \multicolumn{1}{c|}{28.72} & \multicolumn{1}{c}{\textbf{29.27}} \\ \cline{2-19}

		& \multicolumn{1}{c|}{\multirow{3}{*}{\textbf{\rotatebox{90}{SUM}}}} & \textbf{MultiNews} & \multicolumn{1}{c|}{20.59} & \multicolumn{1}{c}{19.78} & \multicolumn{1}{c}{18.12} & \multicolumn{1}{c}{19.91} & \multicolumn{1}{c}{20.49} & \multicolumn{1}{c}{20.52} & \multicolumn{1}{c|}{\textbf{20.79}} & \multicolumn{1}{c|}{\color{black}{20.66}} & \multicolumn{1}{c|}{18.37} & \multicolumn{1}{c}{15.54} & \multicolumn{1}{c}{14.28} & \multicolumn{1}{c}{17.53} & \multicolumn{1}{c}{18.09} & \multicolumn{1}{c}{18.26} &  \multicolumn{1}{c|}{18.19} & \multicolumn{1}{c}{\textbf{18.44}} \\ 
		& \multicolumn{1}{c|}{}                            & \textbf{GovReport} & \multicolumn{1}{c|}{24.08} & \multicolumn{1}{c}{18.39} & \multicolumn{1}{c}{15.92} & \multicolumn{1}{c}{18.09} & \multicolumn{1}{c}{23.50} & \multicolumn{1}{c}{24.01} & \multicolumn{1}{c|}{23.74} & \multicolumn{1}{c|}{\textbf{22.88}} & \multicolumn{1}{c|}{20.54} & \multicolumn{1}{c}{16.24} & \multicolumn{1}{c}{15.25} & \multicolumn{1}{c}{15.99} &  \multicolumn{1}{c}{21.00} & \multicolumn{1}{c}{\textbf{21.55}} &\multicolumn{1}{c|}{20.70} & \multicolumn{1}{c}{\color{black}{20.68}} \\
		& \multicolumn{1}{c|}{}                            & \textbf{QMSum} & \multicolumn{1}{c|}{20.51} & \multicolumn{1}{c}{17.90} & \multicolumn{1}{c}{17.84} & \multicolumn{1}{c}{15.08} & \multicolumn{1}{c}{17.62} & \multicolumn{1}{c}{17.93} &\multicolumn{1}{c|}{20.20} &  \multicolumn{1}{c|}{\color{black}{\textbf{20.41}}} & \multicolumn{1}{c|}{20.04} & \multicolumn{1}{c}{17.79} & \multicolumn{1}{c}{17.29} & \multicolumn{1}{c}{15.06} &  \multicolumn{1}{c}{18.57} & \multicolumn{1}{c}{18.67} &\multicolumn{1}{c|}{19.46} & \multicolumn{1}{c}{\textbf{19.81}}  \\ \cline{2-19}

		& \multicolumn{1}{c|}{\multirow{2}{*}{\textbf{\rotatebox{90}{FS}}}} & \textbf{TREC} & \multicolumn{1}{c|}{50.00} & \multicolumn{1}{c}{50.00} & \multicolumn{1}{c}{48.74} & \multicolumn{1}{c}{50.25} & \multicolumn{1}{c}{50.00} & \multicolumn{1}{c}{50.76}& \multicolumn{1}{c|}{52.27}  & \multicolumn{1}{c|}{\textbf{56.03}} & \multicolumn{1}{c|}{64.07} & \multicolumn{1}{c}{62.06} & \multicolumn{1}{c}{60.81} & \multicolumn{1}{c}{63.32} & \multicolumn{1}{c}{63.82} & \multicolumn{1}{c}{40.71} & \multicolumn{1}{c|}{64.07} & \multicolumn{1}{c}{\color{black}{\textbf{65.33}}} \\ 
		& \multicolumn{1}{c|}{}                            & \textbf{SAMSUM} & \multicolumn{1}{c|}{10.62} & \multicolumn{1}{c}{12.03} & \multicolumn{1}{c}{11.83} & \multicolumn{1}{c}{13.34} &\multicolumn{1}{c}{11.10} & \multicolumn{1}{c}{10.92} & \multicolumn{1}{c|}{10.62} &  \multicolumn{1}{c|}{\textbf{17.43}} & \multicolumn{1}{c|}{12.38} & \multicolumn{1}{c}{12.55} & \multicolumn{1}{c}{13.28} & \multicolumn{1}{c}{13.83} & \multicolumn{1}{c}{\textbf{15.65}} & \multicolumn{1}{c}{12.95} & \multicolumn{1}{c|}{12.93} & \multicolumn{1}{c}{\color{black}{12.40}}  \\ \hline\hline

		\multirow{11}{*}{\rotatebox{90}{\textbf{End}}} & \multicolumn{1}{c|}{\multirow{6}{*}{\textbf{\rotatebox{90}{QA}}}} & \textbf{MFQA-en}& \multicolumn{1}{c|}{50.93} & \multicolumn{1}{c}{47.93} & \multicolumn{1}{c}{48.38} & \multicolumn{1}{c}{32.52} &  \multicolumn{1}{c}{28.62} & \multicolumn{1}{c}{51.40} &\multicolumn{1}{c|}{49.67} & \multicolumn{1}{c|}{\color{black}{\textbf{51.69}}} & \multicolumn{1}{c|}{48.82} & \multicolumn{1}{c}{47.57} & \multicolumn{1}{c}{47.24} & \multicolumn{1}{c}{29.28} & \multicolumn{1}{c}{27.66} & \multicolumn{1}{c}{47.94}& \multicolumn{1}{c|}{\textbf{49.18}}  & \multicolumn{1}{c}{\color{black}{48.67}} \\
		& \multicolumn{1}{c|}{}                            & \textbf{2WikiMQA}& \multicolumn{1}{c|}{42.43} & \multicolumn{1}{c}{42.34} & \multicolumn{1}{c}{41.96} & \multicolumn{1}{c}{37.20} & \multicolumn{1}{c}{25.19} & \multicolumn{1}{c}{39.54} & \multicolumn{1}{c|}{41.75} & \multicolumn{1}{c|}{\textbf{44.05}} & \multicolumn{1}{c|}{42.70} & \multicolumn{1}{c}{\textbf{42.25}} & \multicolumn{1}{c}{41.24} & \multicolumn{1}{c}{32.89} &  \multicolumn{1}{c}{26.50} & \multicolumn{1}{c}{37.54} &\multicolumn{1}{c|}{41.34} & \multicolumn{1}{c}{\color{black}{42.24}} \\
		& \multicolumn{1}{c|}{}                            & \textbf{Musique} & \multicolumn{1}{c|}{29.39} & \multicolumn{1}{c}{28.07} & \multicolumn{1}{c}{29.01} & \multicolumn{1}{c}{20.96} & \multicolumn{1}{c}{7.80} & \multicolumn{1}{c}{26.44} &\multicolumn{1}{c|}{23.56} &  \multicolumn{1}{c|}{\textbf{31.60}} & \multicolumn{1}{c|}{24.18} & \multicolumn{1}{c}{23.08} & \multicolumn{1}{c}{22.39} & \multicolumn{1}{c}{11.82} & \multicolumn{1}{c}{8.68} & \multicolumn{1}{c}{17.65} & \multicolumn{1}{c|}{24.34} & \multicolumn{1}{c}{\color{black}{\textbf{24.82}}} \\
		& \multicolumn{1}{c|}{}                            & \textbf{HotpotQA}  & \multicolumn{1}{c|}{53.62} & \multicolumn{1}{c}{52.38} & \multicolumn{1}{c}{53.59} & \multicolumn{1}{c}{43.83}& \multicolumn{1}{c}{25.04} & \multicolumn{1}{c}{51.71}  & \multicolumn{1}{c|}{51.97} & \multicolumn{1}{c|}{\textbf{55.05}} & \multicolumn{1}{c|}{53.09} & \multicolumn{1}{c}{51.03} & \multicolumn{1}{c}{50.71} & \multicolumn{1}{c}{35.64} & \multicolumn{1}{c}{24.63} & \multicolumn{1}{c}{43.67} &  \multicolumn{1}{c|}{53.01} &\multicolumn{1}{c}{\color{black}{\textbf{53.13}}} \\
		& \multicolumn{1}{c|}{}                            & \textbf{NrtvQA} & \multicolumn{1}{c|}{25.76} & \multicolumn{1}{c}{25.50} & \multicolumn{1}{c}{24.58} & \multicolumn{1}{c}{19.28} & \multicolumn{1}{c}{14.19} & \multicolumn{1}{c}{23.90} & \multicolumn{1}{c|}{23.87} & \multicolumn{1}{c|}{\textbf{25.87}} & \multicolumn{1}{c|}{19.07} & \multicolumn{1}{c}{17.90} & \multicolumn{1}{c}{16.75} & \multicolumn{1}{c}{14.08} &  \multicolumn{1}{c}{11.41} & \multicolumn{1}{c}{14.39} &\multicolumn{1}{c|}{18.37} & \multicolumn{1}{c}{\textbf{19.86}} \\
		& \multicolumn{1}{c|}{}                            & \textbf{Qasper} & \multicolumn{1}{c|}{37.97} & \multicolumn{1}{c}{35.95} & \multicolumn{1}{c}{33.98} & \multicolumn{1}{c}{28.01} & \multicolumn{1}{c}{27.37} & \multicolumn{1}{c}{\textbf{38.67}} & \multicolumn{1}{c|}{36.50} & \multicolumn{1}{c|}{\color{black}{38.27}} & \multicolumn{1}{c|}{33.55} & \multicolumn{1}{c}{31.82} & \multicolumn{1}{c}{30.21} & \multicolumn{1}{c}{24.28} &  \multicolumn{1}{c}{25.15} & \multicolumn{1}{c}{33.72} & \multicolumn{1}{c|}{\textbf{34.18}} &\multicolumn{1}{c}{\color{black}{33.18}}  \\ \cline{2-19}

		& \multicolumn{1}{c|}{\multirow{3}{*}{\textbf{\rotatebox{90}{SUM}}}} & \textbf{MultiNews}& \multicolumn{1}{c|}{20.59} & \multicolumn{1}{c}{20.03} & \multicolumn{1}{c}{18.87} & \multicolumn{1}{c}{19.97} & \multicolumn{1}{c}{20.16} & \multicolumn{1}{c}{20.40} & \multicolumn{1}{c|}{20.49} & \multicolumn{1}{c|}{\textbf{20.45}} & \multicolumn{1}{c|}{18.31} & \multicolumn{1}{c}{16.61} & \multicolumn{1}{c}{15.32} & \multicolumn{1}{c}{17.74}& \multicolumn{1}{c}{17.99} & \multicolumn{1}{c}{18.01}  & \multicolumn{1}{c|}{18.36} & \multicolumn{1}{c}{\color{black}{\textbf{18.40}}} \\ 
		& \multicolumn{1}{c|}{}                            & \textbf{GovReport} & \multicolumn{1}{c|}{23.90} & \multicolumn{1}{c}{20.10} & \multicolumn{1}{c}{18.54} & \multicolumn{1}{c}{18.04} &  \multicolumn{1}{c}{23.17} & \multicolumn{1}{c}{23.35} & \multicolumn{1}{c|}{23.92} &\multicolumn{1}{c|}{\color{black}{\textbf{24.27}}} & \multicolumn{1}{c|}{21.21} & \multicolumn{1}{c}{18.06} & \multicolumn{1}{c}{17.09} & \multicolumn{1}{c}{16.95} &\multicolumn{1}{c}{21.53} & \multicolumn{1}{c}{\textbf{21.82}} & \multicolumn{1}{c|}{21.28} &  \multicolumn{1}{c}{\color{black}{21.10}} \\
		& \multicolumn{1}{c|}{}                            & \textbf{QMSum} & \multicolumn{1}{c|}{22.69} & \multicolumn{1}{c}{22.21} & \multicolumn{1}{c}{22.12} & \multicolumn{1}{c}{19.67} &  \multicolumn{1}{c}{18.59} & \multicolumn{1}{c}{22.28} &\multicolumn{1}{c|}{22.27} & \multicolumn{1}{c|}{\textbf{22.82}} & \multicolumn{1}{c|}{21.17} & \multicolumn{1}{c}{20.15} & \multicolumn{1}{c}{20.03} & \multicolumn{1}{c}{18.13} &  \multicolumn{1}{c}{18.07} & \multicolumn{1}{c}{20.33} &\multicolumn{1}{c|}{\textbf{21.04}} & \multicolumn{1}{c}{\color{black}{20.95}} \\ \cline{2-19}

		& \multicolumn{1}{c|}{\multirow{2}{*}{\textbf{\rotatebox{90}{FS}}}} & \textbf{TREC} & \multicolumn{1}{c|}{59.05} & \multicolumn{1}{c}{58.54} & \multicolumn{1}{c}{58.54} & \multicolumn{1}{c}{58.54} &  \multicolumn{1}{c}{52.51} & \multicolumn{1}{c}{59.55} &\multicolumn{1}{c|}{59.05} & \multicolumn{1}{c|}{\textbf{60.31}} & \multicolumn{1}{c|}{68.34} & \multicolumn{1}{c}{66.08} & \multicolumn{1}{c}{66.59} & \multicolumn{1}{c}{67.84} & \multicolumn{1}{c}{63.82} & \multicolumn{1}{c}{67.59} &\multicolumn{1}{c|}{\textbf{68.85}} & \multicolumn{1}{c}{\color{black}{68.34}} \\
		& \multicolumn{1}{c|}{}                            & \textbf{SAMSUM} & \multicolumn{1}{c|}{18.78} & \multicolumn{1}{c}{\textbf{23.88}} & \multicolumn{1}{c}{20.63} & \multicolumn{1}{c}{19.84} & \multicolumn{1}{c}{18.45} & \multicolumn{1}{c}{17.75} & \multicolumn{1}{c|}{17.62} & \multicolumn{1}{c|}{\color{black}{23.53}} & \multicolumn{1}{c|}{39.46} & \multicolumn{1}{c}{39.58} & \multicolumn{1}{c}{38.77} & \multicolumn{1}{c}{38.71} &  \multicolumn{1}{c}{39.13} & \multicolumn{1}{c}{39.20} & \multicolumn{1}{c|}{39.57} &\multicolumn{1}{c}{\color{black}{\textbf{39.85}}} \\ \hline
	\end{tabular}
\end{table*}

\subsection{Main Experiments}\label{ssec:main_results}
\noindent\textbf{Effectiveness Evaluation.}
To evaluate LoopServe and baselines, we conduct experiments on the proposed  11 multi-turn long-context datasets across three tasks: QA, Summarization (SUM), and Few-shot Learning (FS).
 For each dataset, we compare LoopServe with six state-of-the-art KV cache acceleration baselines and two base LLMs, using F1, Rouge-L, or Accuracy as appropriate.
As shown in Table~\ref{tab:effectiveness}, LoopServe achieves the best or comparable results across most datasets and query positions. 
Notably, LoopServe maintains strong performance regardless of query location, while baselines like SnapKV and AdaKV perform well only when the query is at the end. This highlights their reliance on positional heuristics, which limits generalization. In contrast, LoopServe’s adaptive approach consistently yields higher accuracy and quality, even as context length increases. These gains hold for both Llama-3.1 and Qwen2.5, showing LoopServe generalizes well across LLMs.



\noindent\textbf{Efficiency Evaluation.}
Beyond effectiveness, we also assess LoopServe’s generation efficiency. As shown in Figure~\ref{fig:ablation_study_param_s_alpha}~(a), LoopServe delivers the highest efficiency.
This is achieved through efficient online sparsification, which selects only the most critical attention components, and adaptive KV compression, which maintains a compact, relevant cache. Together, these mechanisms reduce computation and memory usage, enabling fast and high-quality generation.

\noindent\textbf{Ablation Study.}
We explore LoopServe-D (progressive KV compression only) and LoopServe-P (online prefilling sparsification only) on three datasets (MF, 2WM, Qsp) using Llama and Qwen. 
As shown in Figure~\ref{fig:ablation_study_param_s_alpha}~(b) and Figure~\ref{ablation-study-in-qwen} in Appendix, 
LoopServe achieves the best performance, indicating both components are essential and complementary. This advantage holds across tasks, datasets, query positions, and model architectures.
The ablation study reveals that these two components are complementary: while each addresses a different bottleneck in LLM inference, their combination ensures robust adaptation to diverse input patterns and maximizes both efficiency and accuracy.

\subsection{Parameter Sensitivity}\label{ssec:param}
We analyze all hyperparameters in LoopServe: attention sparsity threshold $\alpha$ in online prefilling, token budget $B$, and decoding interval $n_d$ in progressive KV compression.
Due to space limit, the analysis of the parameter $n_d$ is presented in Appendix~\ref{sssec:n_d_param}.

\noindent\textbf{Threshold $\alpha$ in ~\Eqref{eq:prefilling_obj}.}
The parameter $\alpha$ controls how much total attention weight is preserved in prefilling. Higher  $\alpha$ keeps more information but increases computation; lower $\alpha$ boosts efficiency but may lose context.
We evaluate LoopServe on 2WikiMQA and Qasper, with Llama and Qwen backbones, across questions at the beginning (-B), middle (-M), and end (-E) positions, varying $\alpha \in \{0.980, 0.985, 0.990, 0.995, 1.00\}$. As shown in Figure~\ref{fig:ablation_study_param_s_alpha}~(c) and Figure~\ref{fig:param_s_B_and_param_s_nd}~(a) in Appendix, LoopServe get the best accuracy and efficient for $\alpha$ between 0.99 and 1.00. Setting $\alpha$ too low hurts quality, while values close to 1.00 reduce efficiency gains. Overall, LoopServe is not  overly sensitive to $\alpha$ within this range, allowing users to balance speed and quality.


\noindent\textbf{Budget $B$.}
The token selection budget $B$ in LoopServe's progressive KV compression controls the trade-off between efficiency and output quality. We evaluate this on MultiFieldQA and Qasper with queries at the beginning, middle, and end. As shown in Figure~\ref{fig:ablation_study_param_s_alpha}~(d) and Figure~\ref{fig:param_s_B_and_param_s_nd}~(b) in Appendix, increasing $B$ improves accuracy by preserving more relevant tokens, but gains are limited beyond 1024 tokens while computation and memory costs rise. Smaller budgets (256 or 512) reduce accuracy, especially for queries at the beginning or middle, as important tokens may be missed. End-position queries are less affected since key tokens are already cached. Overall, a budget of 1024–2048 tokens offers the best balance of performance and efficiency across all query positions.

\begin{figure}[]
	\centering 	
	\vspace{-3em}
	\subfloat[{Generation latency.}]		
	{\centering\includegraphics[width=0.23\linewidth]{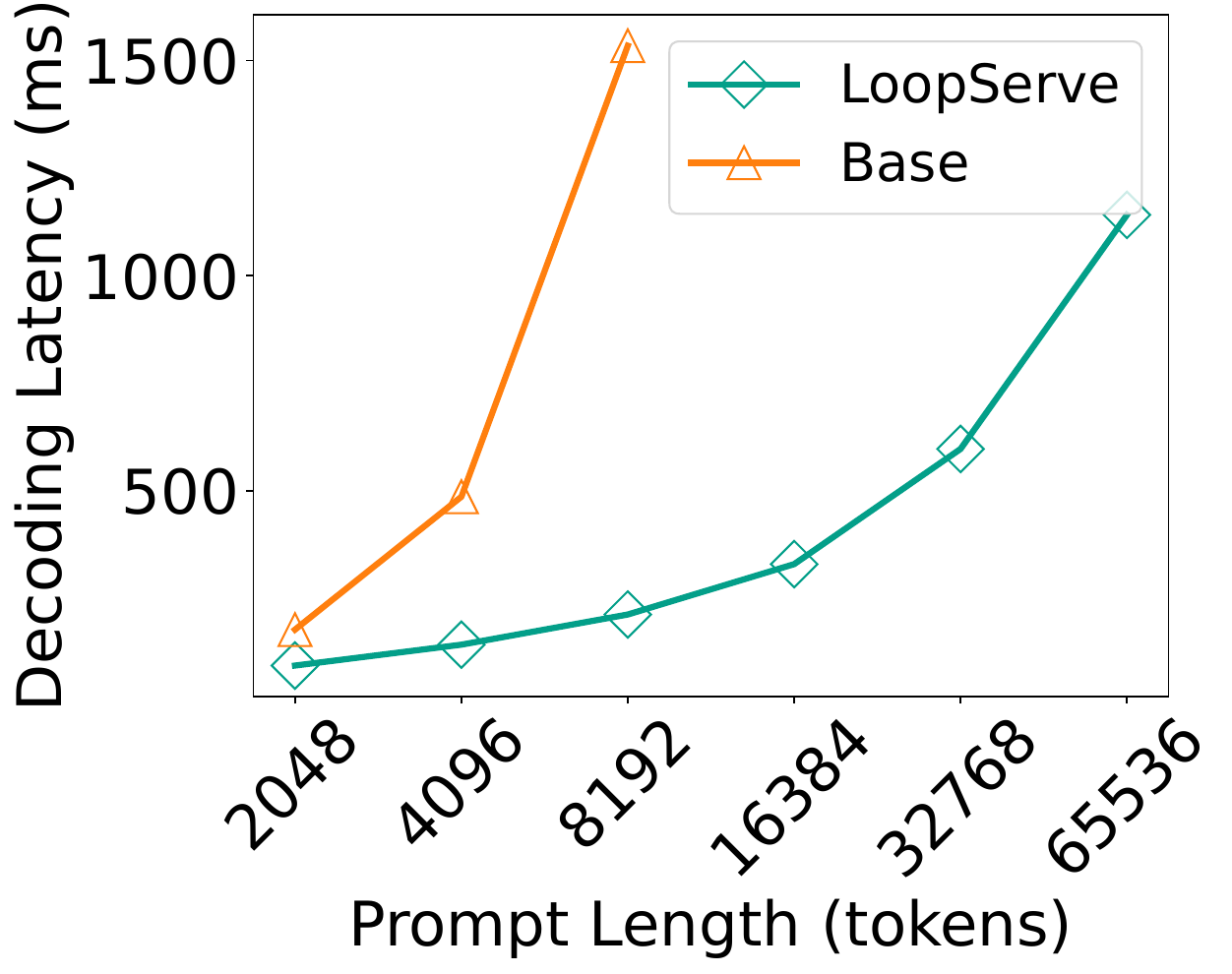}}
	\hfill
	\subfloat[Llama-3.1-8B-Instruct. ]		
	{\centering\includegraphics[width=0.26\linewidth]{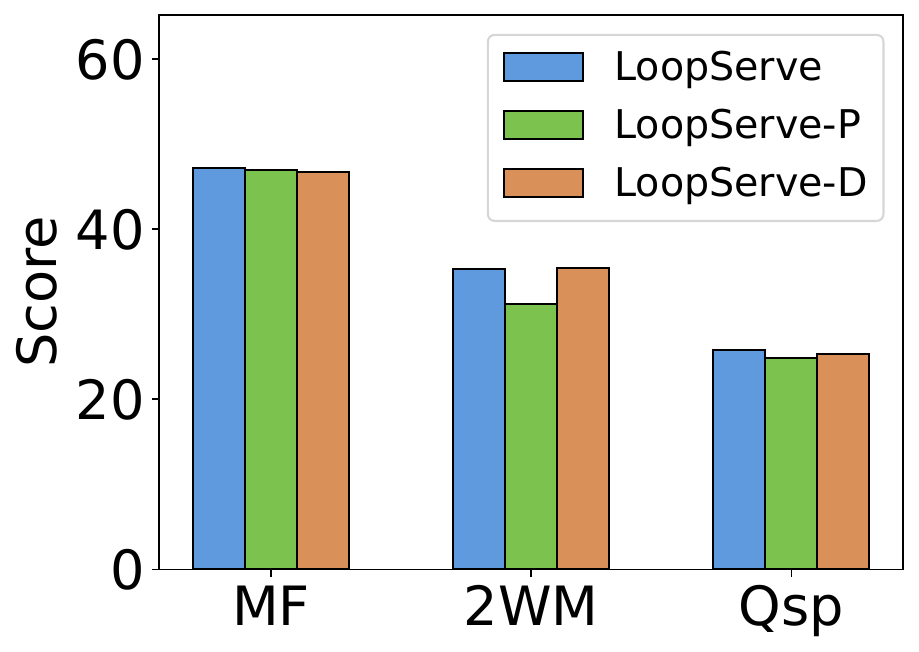}}	
	\hfill
	\subfloat[Llama-3.1-8B-Instruct.]		
	{\centering\includegraphics[width=0.245\linewidth]{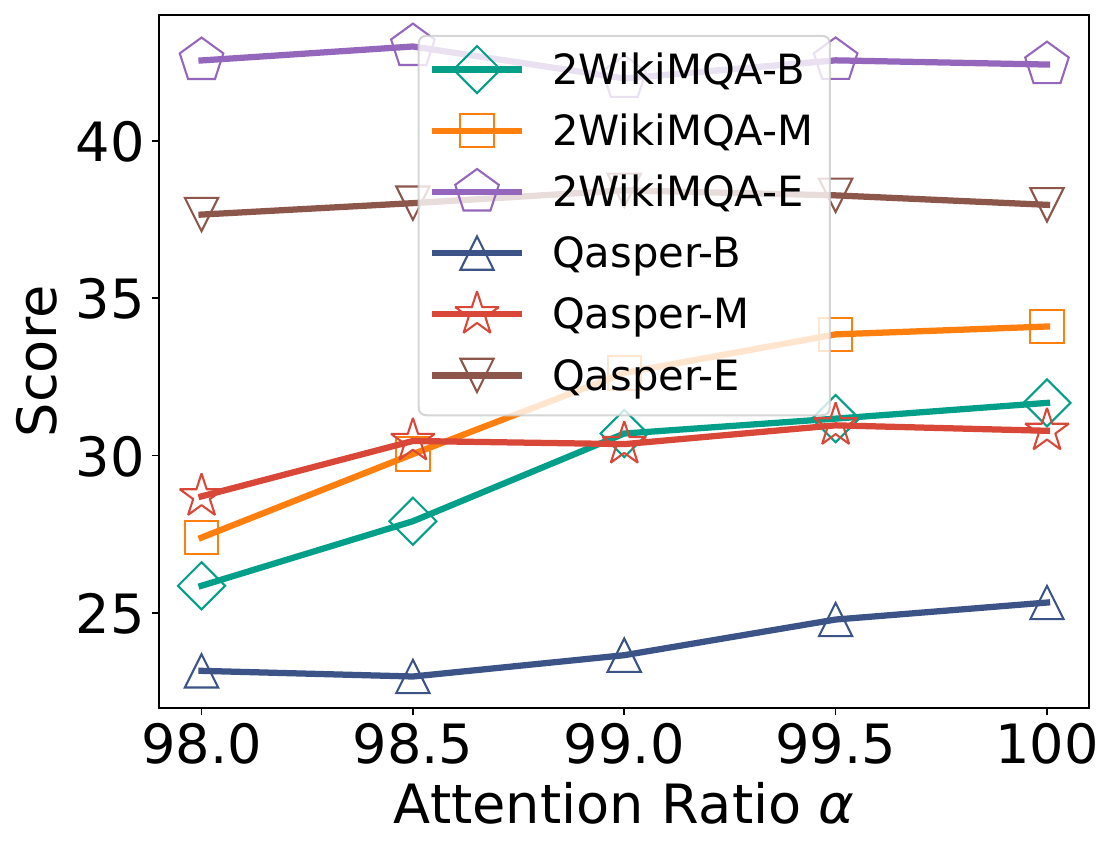}}	
	\hfill
	\subfloat[MultiFieldQA. ]		
	{\centering\includegraphics[width=0.26\linewidth]{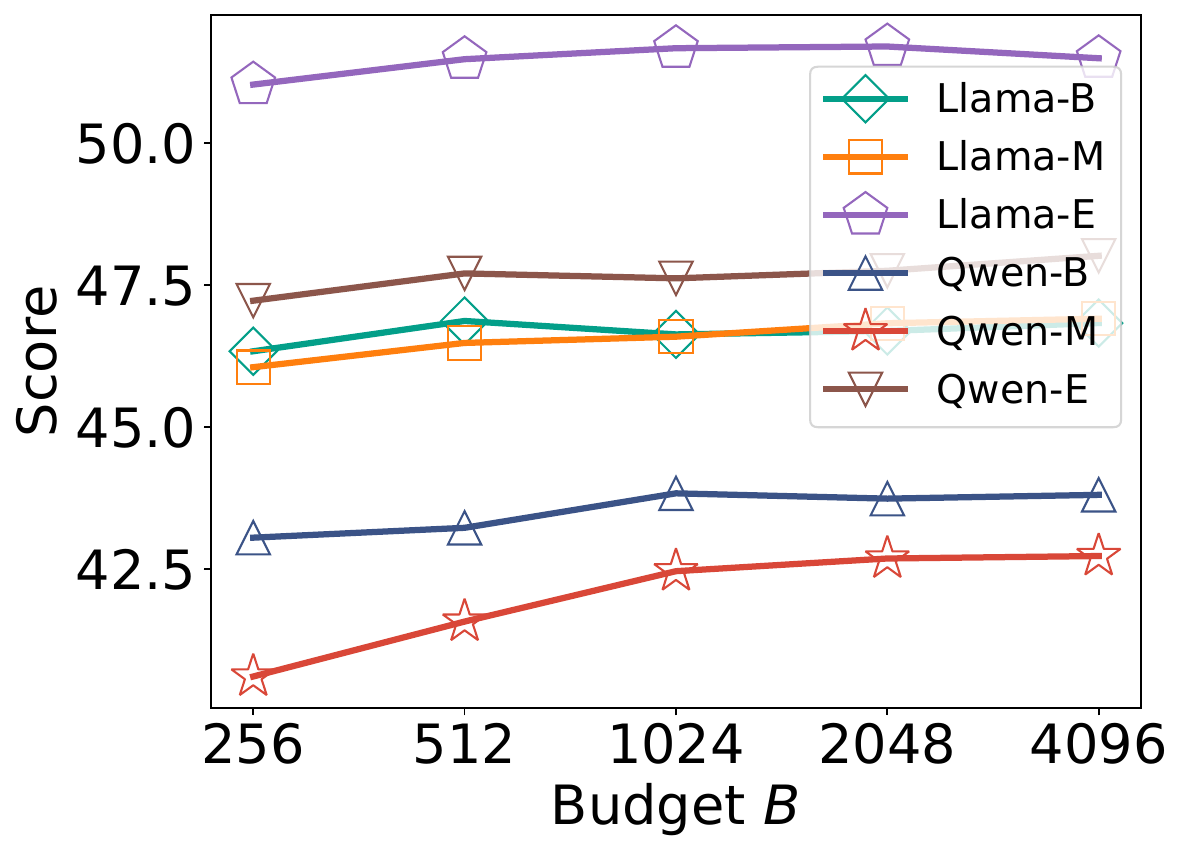}}	
%
	\vspace{-1em}
	\caption{Efficiency in (a), ablation study in (b), and parameter Sensitivity  in (c) and (d).} 
\label{fig:ablation_study_param_s_alpha}
\end{figure}

\vspace{-1em}
\section{Conclusion}\label{sec:conclusion} 
\vspace{-1em}
In this paper, we propose LoopServe, an adaptive dual-phase LLM inference acceleration system designed for realistic multi-turn dialogues. By combining online attention sparsification and progressive KV compression, LoopServe addresses the limitations of static acceleration methods and adapts efficiently to dynamic conversational patterns. Our experiments on diverse, multi-turn benchmarks show that LoopServe significantly improves both inference speed and output quality compared to existing baselines, regardless of query position. This work provides a practical solution for efficient and effective LLM deployment in real-world dialogue scenarios.

{\tiny {\tiny }}
\bibliography{kvcache}
\bibliographystyle{iclr2026_conference}

\appendix
\clearpage
\appendix

\section{Appendix}

\subsection{Important Notations Table}

\begin{table}[h]
	\centering
	\setlength\tabcolsep{5pt}
	\renewcommand{\arraystretch}{1.05}
	\caption{Summary of important notations.}
	\label{tab:notation}
	\begin{tabular}{c|l}
		\hline
		\textbf{Symbol} & \textbf{Definition} \\
		\hline
		$X_i$ & Input sequence of tokens \\ \hline
		
		$X_{i,j}$ & The $j$-turn input of $X_i$ \\ \hline
		
		$Y_i$ & Output sequence of tokens \\ \hline
		
		$y_{i,j}$ & The $j$-turn output of $X_i$ \\ \hline

		$n_i, n_{i,j}$ & Length of  input sequence $X_i$ and $X_{i,j}$ \\ \hline
		
		$m_i, m_{i,j}$ & Length of  output sequence $Y_i$ and $y_{i,j}$ \\ \hline
		
		$M_\theta$ & LLM model \\ \hline
		
		$n_h$ & The total number of attention head of $M_\theta$ \\ \hline

		$\mathbf{Q}^k_i, \mathbf{K}^k_i, \mathbf{V}^k_i$ & Query, Key, and Value matrices \\  \hline

		$\mathbf{A}^k_i$  & The $k$-th attention head $X_i$ \\ \hline

		$\mathcal{S}^k_{i}$,	$\mathcal{V}^k_{i}$ & Slash and vertical lines of  head $\mathbf{A}^k_i$ \\ \hline
		
		$\mathcal{\hat{S}}^k_{i},\mathcal{\hat{V}}^k_{i}$ &  Selected slash lines and vertical lines  \\ \hline
		
		$n_d$  & Decoding interval  \\ \hline
		
		$B$  &  Budget for input tokens  \\ \hline
		
		$\hat{X}^k_i$  &  Selected important  tokens for attention head $k$  \\  
		
		\hline
	\end{tabular}
\end{table}

Table~\ref{tab:notation} provides detailed definitions of important notations appearing in this paper.

\subsection{Summary Table of KV-based approaches}

\begin{table}[h]
	\centering
	\setlength{\tabcolsep}{2pt}
	\caption{LLM acceleration model comparisons, following~\citet{li2025scbenchkvcachecentricanalysis}. P and D denote whether the model has optimization in the Prefilling and Decoding phases, respectively. $n$ is the token size of the input, $m$ is the generation token size, and $c$ and $k$ are constants with $c, k \ll n$ and $c, k \ll m$.}
	\label{tab:long_context_methods}
	\begin{tabular}{l|cc|ccc} 
		\toprule
		\textbf{Methods} & \textbf{\textcolor{black}{P}} & \textbf{\textcolor{black}{D}} & \textbf{KV  Size} & \textbf{Prefilling } & \textbf{Decoding } \\
		\midrule
		
		\textbf{LLMLingua}~\cite{pan2024llmlingua}   & \checkmark & $\times$ & $O(\alpha n)$ & $O(\alpha^2 n^2)$ & $O(\alpha nm)$ \\
		
		\textbf{A-shape}~\cite{xiao2024efficient}  & \checkmark & $\times$ & $O(n)$ & $O(kn)$ & $O(nm)$ \\
		
		\textbf{Tri-shape}~\cite{li2025scbenchkvcachecentricanalysis}    & \checkmark & $\times$ & $O(n)$ & $O(kn)$ & $O(nm)$ \\
		
		\textbf{MInference}~\cite{jiang2024minference} & \checkmark & $\times$ & $O(n)$ & $O(kn)$ & $O(nm)$ \\
		
		\textbf{SLLM}~\cite{xiao2024efficient}  & $\times$ & \checkmark & $O(k)$ & $O(n^2)$ & $O(km)$ \\
		
		\textbf{SnapKV}~\cite{li2024snapkv}  & $\times$ & \checkmark & $O(k)$ & $O(n^2)$ & $O(km)$ \\
		
		\textbf{AdaKV}~\cite{feng2024ada}  & $\times$ & \checkmark & $O(k)$ & $O(n^2)$ & $O(km)$ \\

		\midrule
		\textbf{LoopServe} &\checkmark & \checkmark & $O(k)$ & $O(kn)$ & $O(k(m-c) + nc)$ \\
		\bottomrule
	\end{tabular}
\end{table}

Table~\ref{tab:long_context_methods} summarizes the time complexity for each KV-based approach.

\subsection{Supplementary Figure for Motivational Observation 2}

\begin{figure}[h]
	\centering
	\begin{minipage}{0.45\textwidth}
		\centering
		\includegraphics[width=0.8\textwidth]{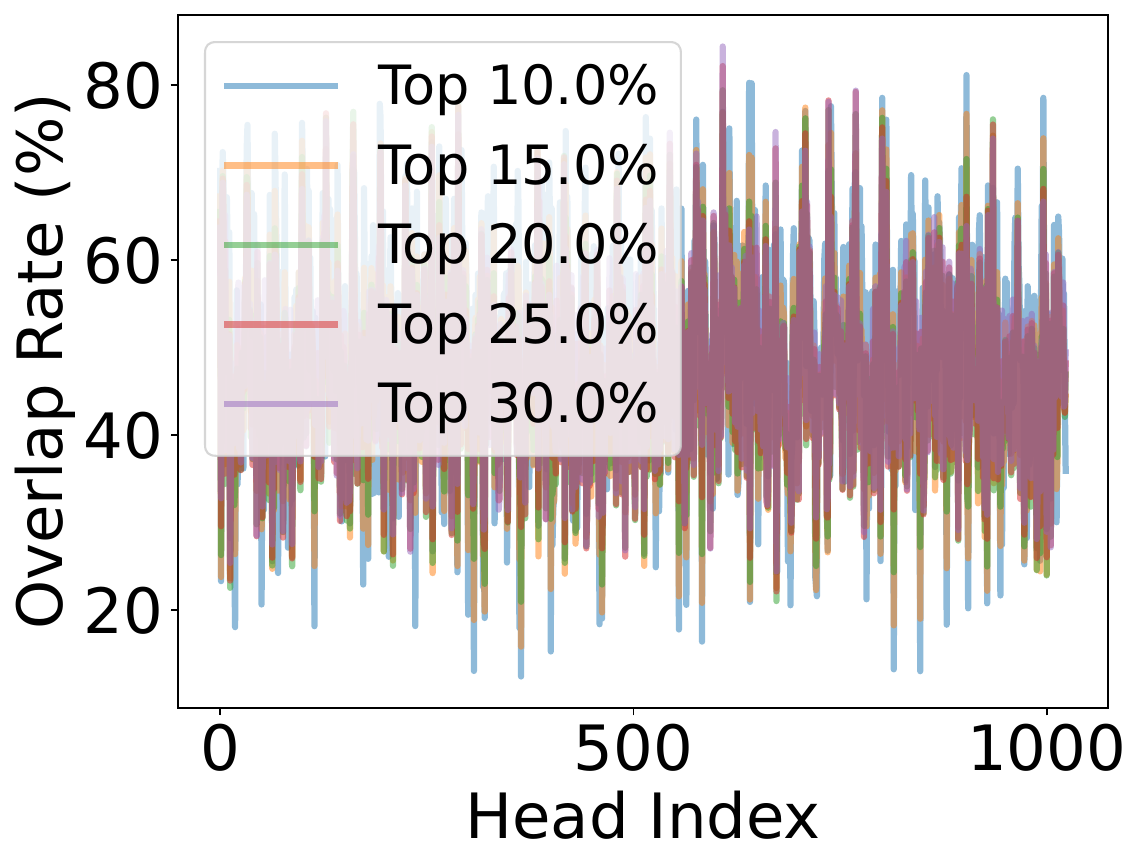}
		\caption{{Overlap rate of each head regarding different inputs of Mistral-7B-Instruct-v0.3
				}}
		\label{fig:mv2_diff}
	\end{minipage}
	\hfill
	\begin{minipage}{0.45\textwidth}
		\centering
		\includegraphics[width=0.9\textwidth]{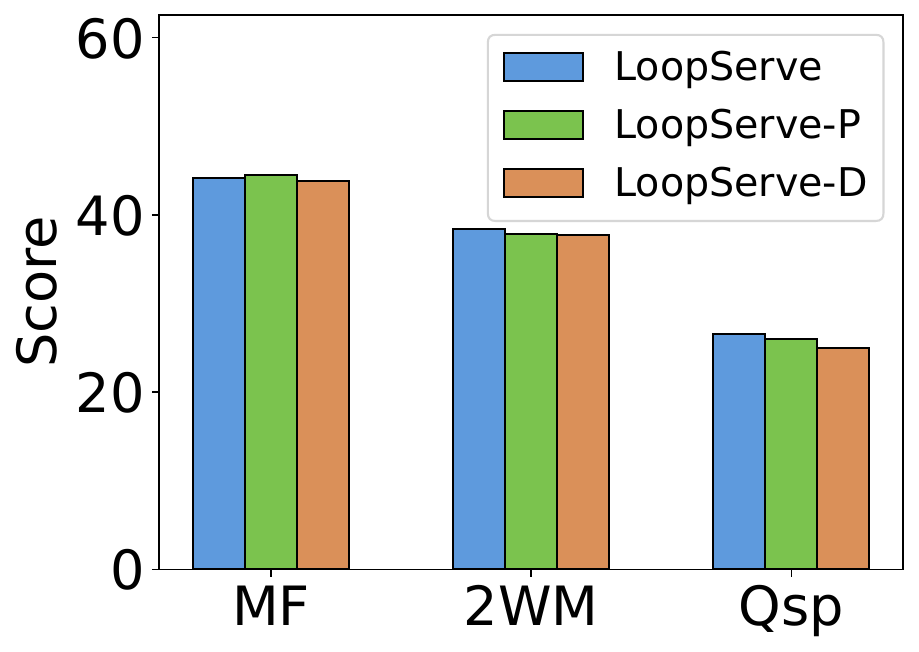}
		\caption{{Ablation Study of LoopServe on Qwen2.5-7B-Instruct.}}
		\label{ablation-study-in-qwen}
	\end{minipage}
\end{figure}

Figure~\ref{fig:mv2_diff} shows  that for most heads, the overlap of Mistral-7B-Instruct-v0.3 remains below 0.5.
Please refer to the detailed analysis in Section~\ref{ssec:kp1}} motivational experiment 2.

\subsection{Supplementary Results for Ablation Study}

Figure~\ref{ablation-study-in-qwen} shows that LoopServe applied on Qwen2.5-7B-Instruct achieves the best performance, indicating the significance of both components.
Please refer to the detailed analysis in ablation study in Section~\ref{ssec:main_results}.

\subsection{Long-context Multi-turn LongBench}\label{appx:ssec:longbench}
Recently, various long-context benchmarks, such as NumericBench~\cite{li2025exposingnumeracygapsbenchmark}, LongBench~\cite{bai2024longbench}, and LongEval~\cite{longchat2023} have been proposed to evaluate LLMs. However, these benchmarks have two main limitations: (1) They assume user queries always appear at the end of the input, which does not reflect real-world scenarios with queries at arbitrary positions. This bias favors KV-compression methods optimized for end-positioned queries, limiting their generalizability. (2) Most benchmarks are single-turn, overlooking the multi-turn dependencies crucial for realistic conversations~\cite{li2025scbenchkvcachecentricanalysis}. 

To overcome these issues, we propose a multi-turn benchmark spanning 11 datasets with diverse query positions and interaction patterns, enabling more realistic long-context LLM evaluation.

\subsubsection{The Design of Multi-turn LongBench}\label{ssec:design}
We represent each $m$-turn long-context data instance in our dataset using a structured format. Specifically, each data instance $\mathcal{I}_i$ consists of $m$ turns, where each turn contains a triplet of context, question, and answer. The complete dataset can be formally denoted as:
$$\mathcal{D}= \{\mathcal{I}_i=[(C_{i,1}, q_{i,1}, a_{i,1}), (C_{i,2}, q_{i,2}, a_{i,2}), \dots, (C_{i,m}, q_{i,m}, a_{i,m})]\}_{i=1}^{|\mathcal{D}|},$$
where $C_{i,j}$ is the context at the $j$-th turn of the instance $\mathcal{I}_i$, which can be empty, $q_{i,j}$ is the corresponding user question, and $a_{i,j}$ denotes the generated answer of the LLMs for $q_{i,j}$.
We design diverse formats for each multi-turn long-context data instance as follows:


\begin{itemize}[leftmargin=*]
	\item \textbf{Diverse Query Positions:}
	For the $j$-th turn, given a context $C_{i,j} = \{C^1_{i,j}, C^2_{i,j}, \dots, C^{p}_{i,j}\}$ consisting of $p$ distinct paragraphs (e.g., segments), 
	the query $q_{i,j}$ can be positioned at various locations within $C_{i,j}$. Specifically, it can appear at the beginning of $C_{i,j}$, at the end of $C_{i,j}$, or between two segments $C^{k}_{i,j}$ and $C^{k+1}_{i,j}$.
	Such a way addresses the limitation of existing methods, which only place the query at the end of the context. This placement may not accurately reflect real-world scenarios.
	
	\item \textbf{Diverse Query Relevance:}
	At the $j$-th turn, the answer $a_{i,j}$ to question $q_{i,j}$ is derived from the contexts $\{C_{i,j'}\}_{j'=1}^j$. 
	In real user scenarios, the context sources for answering $q_{i,j}$ are diverse. 
	Instead of restricting $q_{i,j}$ to rely solely on $C_{i,j}$, we design the answer $a_{i,j}$ to $q_{i,j}$ to come from a subset of contexts $C_{q_i} \subseteq \{C_{i,j'}\}_{j'=1}^j$, with the size of the subset varying as $|C_{q_i}| \in \{1, 2, \dots, j\}$. 
\end{itemize}

\subsubsection{Multi-turn LongBench Generation} \label{ssec:dataset-generation}

\begin{table}[h]
	\centering
	
	\setlength{\tabcolsep}{8pt}
	\renewcommand{\arraystretch}{1.3}
	
	\caption{Multi-turn dataset statistics.}
	\label{multi-turn-dataset-table}
	\begin{tabular}{c|c|c|c|c|c}
		\hline
		\textbf{Type} & \textbf{Dataset} & \textbf{$|D|$} & \textbf{\#Turn} & \textbf{Avg Token} & \textbf{Metric} \\
		\hline
		\multirow{6}{*}{\textbf{QA}} & \textbf{NQA} & 500 & 3 & 30545.54 & F1 \\
		
		& \textbf{Qasper} & 500 & 3 & 5434.41 & F1 \\
		
		& \textbf{MFQA-en} & 500 & 3 & 7279.64 & F1 \\
		
		& \textbf{HotpotQA} & 500 & 3 & 13847.19 & F1 \\
		
		& \textbf{2WikiMQA} & 500 & 3 & 7471.16 & F1 \\
		
		& \textbf{Musique} & 500 & 3 & 16587.56 & F1 \\
		\hline
		\multirow{3}{*}{\textbf{Summary}} & \textbf{MultiNews} & 500 & 2 & 2376.46 & Rouge-L \\
		
		& \textbf{GovRepor}t & 500 & 2 & 9324.43 & Rouge-L \\
		
		& \textbf{QMSum} & 500 & 2 & 12780.29 & Rouge-L \\
		\hline
		\multirow{2}{*}{\begin{tabular}[c]{@{}c@{}} \textbf{Few-shot}\\\textbf{Learning}\end{tabular}} & \textbf{TREC} & 199 & 2 & 2293.99 & Accuracy \\

		& \textbf{SAMSUM} & 199 & 2 & 3113.73 & Accuracy \\
		\hline
	\end{tabular}
\end{table}

Based on the above format, we design multi-turn long-context benchmarks across various categories. Dataset details are in Table~\ref{multi-turn-dataset-table}. Construction methodology follows:

\begin{itemize}[leftmargin=*]
	\item \textbf{Question Answering (QA).} These datasets are derived from the single-document QA and multi-document QA tasks in LongBench~\cite{bai2024longbench,bai2024longbench2}, with each dataset comprising 500 instances. Each instance is structured into three turns. To construct these instances, we randomly select three question-answer pairs from a dataset in LongBench~\cite{bai2024longbench,bai2024longbench2} as the foundation for the three turns. The associated contexts are then systematically modified through splitting and recombination, with additional irrelevant contexts incorporated. This meticulous design ensures that each instance satisfies the requirements for diverse query positions and diverse query relevance, as outlined in Appendix~\ref{ssec:design}.

	\item \textbf{Summarization.} These datasets are derived from the summarization tasks within LongBench~\cite{bai2024longbench,bai2024longbench2}. Each dataset comprises 500 instances, with each instance consisting of two turns. To enhance the diversity of the input, we randomly selected two instances from the original dataset and segmented their original contexts, subsequently recombining them into two turns. This process was carefully designed to ensure compliance with the diverse query relevance requirement outlined in Appendix~\ref{ssec:design}. Finally, we annotated the source paragraphs for traceability and introduced additional noisy contexts to further enrich the complexity and challenge of the dataset.
	
	\item \textbf{Few-shot Learning.} These datasets are derived from the few-shot learning tasks in LongBench~\cite{bai2024longbench,bai2024longbench2}. To fulfill the requirements of diverse query positions and diverse query relevance as outlined in Appendix~\ref{ssec:design}, we exclude instances containing fewer than four examples. For the remaining eligible instances, the examples are segmented and distributed across the first and second turns. The LLM is tasked with generating an initial response based on the examples provided in the first turn and subsequently refining its response in the second turn using the additional examples. Furthermore, the query is strategically positioned at the beginning, middle, and end of the examples to ensure diversity in query placement. To maintain the semantic integrity and structural completeness of the examples, regular expressions are employed for segmentation.
\end{itemize}

\subsection{Proof of Theorem 1}\label{ssec: proof of theorem 1}

\begin{proof}
	The Online Prefilling Sparsification Problem (OPSP) can be proven NP-hard via a reduction from the Set Cover Problem, a well-known NP-hard problem. 
	The Set Cover Problem is defined as follows: given a universe $U = \{u_1, u_2, \dots, u_m\}$, a collection of subsets $\mathcal{P} = \{P_1, P_2, \dots, P_n\}$,  the objective is to find a subset $\mathcal{P}^* \subseteq \mathcal{P}$ such that $\bigcup_{P_i \in \mathcal{P}^*} P_i = U$,
	and the total cost $\sum_{P_i \in \mathcal{P}^*} |P_i|$ is minimized. 
	We map the elements of the Set Cover Problem to the OPSP as follows. 
	Each element $u \in U$ corresponds to an entry in the attention matrix $\mathbf{A}_i^k[\hat{X}_{i,j}]$ that needs to be covered. Each subset $P_i$ in $\mathcal{P}$ corresponds to a slash line $s$ or a vertical line $v$ in OPSP, which covers a subset of entries in the matrix. 
	The cost of selecting subset $P_i$ is mapped to the cost $l_s$ (for slash lines $s$) or $l_v$ (for vertical lines $v$) in OPSP. The Set Cover Problem's requirement to cover all elements in $U$ is equivalent to requiring  $\alpha$   in OPSP. 
	Therefore, if we can solve the OSOP optimally in polynomial time, we can solve the set cover problem in polynomial time.
\end{proof}

\subsection{Algorithms of LoopServe System}\label{ssec:alg}

Algorithm~\ref{alg:framwork}, Algorithm~\ref{alg:prefilling}, and Algorithm~\ref{alg:decoding} of LoopServe System in section~\ref{sec:method} are listed below.

\begin{algorithm}[h]
	
	\KwIn{
		The $m$-turn input $\mathcal{I}_i=\{X_{i,j}\}_{j=1}^m$,  LLM $M_\theta$,  threshold $\alpha$, re-selection interval $n_d$, and budget $B$
	}
	\KwOut{ Answers $\{y_{i,j}\}_{j=1}^m$
	}
	$X_i \gets \emptyset$\\
	\For{$j=1$ \textbf{to} $m$}{
		$X_i = X_i \cup y_{i,j-1} \cup X_{i,j}, \  \hat{X}_{i,j}= [ y_{i,j-1}, X_{i,j}]$\\
		\tcp{Step 1: Parallel Prefilling Line Selection}
		\For{$k=1$ \textbf{to} $n_h$}{
			$\mathcal{\hat{V}}^k_{i,j}, \mathcal{\hat{S}}^k_{i,j}$ = \textsf{PrefillingLineSelection}($M_\theta$, $X_i$, $\hat{X}_{i,j}$,  $\alpha$)
		}

		\tcp{Step 2:  KV Compression for Decoding}
		$\mathcal{L}=\{\mathcal{\hat{V}}^{k}_{i,j'}, \mathcal{\hat{S}}^{k}_{i,j'}\}_{j'=1,k=1}^{j,n_h}$\\
		
		$y_{i,j} =  \textsf{ProgressiveDecoding}(M_\theta, {X}_{i}, 	\mathcal{L}, n_d, B)$\\
		
	}
	\textup{\textbf{Return\ }}  Answers $\{y_{i,j}\}_{j=1}^m$.
	\caption{LoopServe framework overview}
	\label{alg:framwork}
\end{algorithm}

\begin{algorithm}[h]
	 
	\KwIn{
		The input $X_i$ and $\hat{X}_{i,j}$,  $k$-th head of LLM $\mathcal{M}_\theta$,  
		the parameter $\alpha$
	}
	\KwOut{ The selected slash lines $\mathcal{\hat{S}}_{i,j}^k$ and vectical lines  $\mathcal{\hat{V}}_{i,j}^k$ 
	}

	$\tilde{X}_{i,j} = \textsf{RandomSelect}(\hat{X}_{i,j})$\\
	
	\textsf{Compute } Query $\mathbf{\tilde{Q}}^k_{i,j} $ for  $\tilde{X}_{i,j}$\\
	
	\textsf{Compute } Key $\mathbf{{K}}^k_{i} $ for  ${X}_{i}$\\
	
	$\mathbf{{A}}^k_{i}[\tilde{X}_{i,j}] = \textsf{Softmax}\left({\mathbf{\tilde{Q}}^k_{i.j} (\mathbf{K}^k_{i})^\top/\sqrt{d_k}}\right)$\\
	
	
	$\mathcal{S}^k_{i,j}= \textsf{SlashSum}(\mathbf{{A}}^k_{i}[\tilde{X}_{i,j}])$,
	$\mathcal{V}^k_{i,j}= \textsf{VerticalSum}(\mathbf{{A}}^k_{i}[\tilde{X}_{i,j}])$\\
	
	$\mathcal{S}^k_{i,j} \gets \texttt{Desc\_Sort}(\mathcal{S}^k_{i,j})$,
	$\mathcal{V}^k_{i,j} \gets \texttt{Desc\_Sort}(\mathcal{V}^k_{i,j})$\\
	
	$\mathcal{\hat{S}}^k_{i,j} \gets \emptyset$,  $\mathcal{\hat{V}}^k_{i,j}  \gets \emptyset$\\
	$ol_s = 0, ol_v = 0, sum=0$\\
	\While{$sum < \alpha \cdot \textsf{Sum}(\mathbf{{A}}^k_{i}[\tilde{X}_{i,j}] )$}{
		
		$s=\mathcal{{S}}^k_{i,j}[0], v=\mathcal{V}^k_{i,j}[0]$  \\
		
		$\triangle w_s = w_s - ol_v, \triangle w_v = w_v - ol_s$ \\
		\If{$\triangle w_s/(l_s-|\mathcal{\hat{V}}^k_{i,j}|) \ge \triangle w_v/(l_v-|\mathcal{\hat{S}}^k_{i,j}|)$}{
			$\mathcal{\hat{S}}^k_{i,j} =\mathcal{\hat{S}}^k_{i,j} \cup \{s\}$, $ol_s = ol_s + w^{max}_s$\\
			$sum =sum+w_s-ol_v$\\
		}\Else{
			$\mathcal{\hat{V}}^k_{i,j} = \mathcal{\hat{V}}^k_{i,j} \cup \{v\}$, $ol_v = ol_v + w^{max}_v$\\
			$sum =sum+w_v-ol_s$\\
		}
		
	}
	
	\textup{\textbf{Return\ }} $\mathcal{\hat{S}}^k_{i,j}$ and $\mathcal{\hat{V}}^k_{i,j}$.
	\caption{Adaptive Prefilling Sparsification Framework}
	\label{alg:prefilling}
\end{algorithm}

\begin{algorithm}[h]
	 
	\KwIn{
		Input $\mathcal{I}_i$,   LLM $M_\theta$, all selected slash and vertical lines $\{\mathcal{\hat{V}}^{k}_{i,j'}, \mathcal{\hat{S}}^{k}_{i,j'}\}_{j'=1,k=1}^{j,n_h}$,
		re-selection interval $n_d$, and budget $B$
	}
	\KwOut{Answer $y_{i,j}$
	}
	$n_o=0,y_{i,j} =\emptyset$\\
	\While{LLM generation is not finshed}{
		\If{$n_o = 16$ or $(n_o-16)\%n_d=0$}{
			${X}^{obs}_{i}  = X_i[|X_i|-n_d: |X_i|]$ \\
			\lForEach{$k=1$ \textbf{to} $n_h$}{ 
				$ 	\hat{X}^k_i  = \arg\max_{\hat{X}^k_i \subseteq X_i, |\hat{X}^k_i| = B|} \sum_{a \in \hat{X}_i} \sum_{b \in X^{obs}_i} \mathbf{A}^k_i[a][b]$
			}
		}
		$n_o = n_o + 1$ \\
		$x_{n_i+n_o} = \textsf{Decoding}(M_\theta, \{\hat{X}^k_{i}\}_{k=1}^{n_h}, 	\{\mathcal{\hat{V}}^{k}_{i,j'}, \mathcal{\hat{S}}^{k}_{i,j'}\}_{j'=1,k=1}^{j,n_h})$\\
		
		$\hat{X}_{i}= \hat{X}_{i} \cup x_{n_i+n_o}$,
		$  {X}_{i}=  {X}_{i} \cup x_{n_i+n_o} $ \\
		
		$y_{i,j}=y_{i,j} \cup x_{n_i+n_o}$
		
	}  
	
	\textup{\textbf{Return\ }}  Answer $y_{i,j}$
	\caption{Progressive Decoding}
	\label{alg:decoding}
\end{algorithm}

\subsubsection{Time Complexity Analysis of Algorithm~\ref{alg:framwork}}\label{sssec:time_complexity}

It is $O(n_i|\tilde{X}_{i,j}|+n_ilogn_i+n_i)$.
Firstly, it takes $O(n_i|\tilde{X}_{i,j}|)$ to compute the partial attention matrix $\mathbf{{A}}^k_i[\tilde{X}_{i,j}]$.
Then, it takes  $O(n_i|\tilde{X}_{i,j}|)$ to summarize the values for each slash line and vertical line, and takes $O(n_i \log n_i)$ for descending sorts.
Finally,  the greedy selection loop runs in $O(n_i)$.

\subsection{Supplementary Experimental Figure for Parameter Sensitivity}

\begin{figure}[H]
	\centering 	
	\subfloat[Qwen-2.5-7B-Instruct.]		
	{\centering\includegraphics[width=0.26\linewidth]{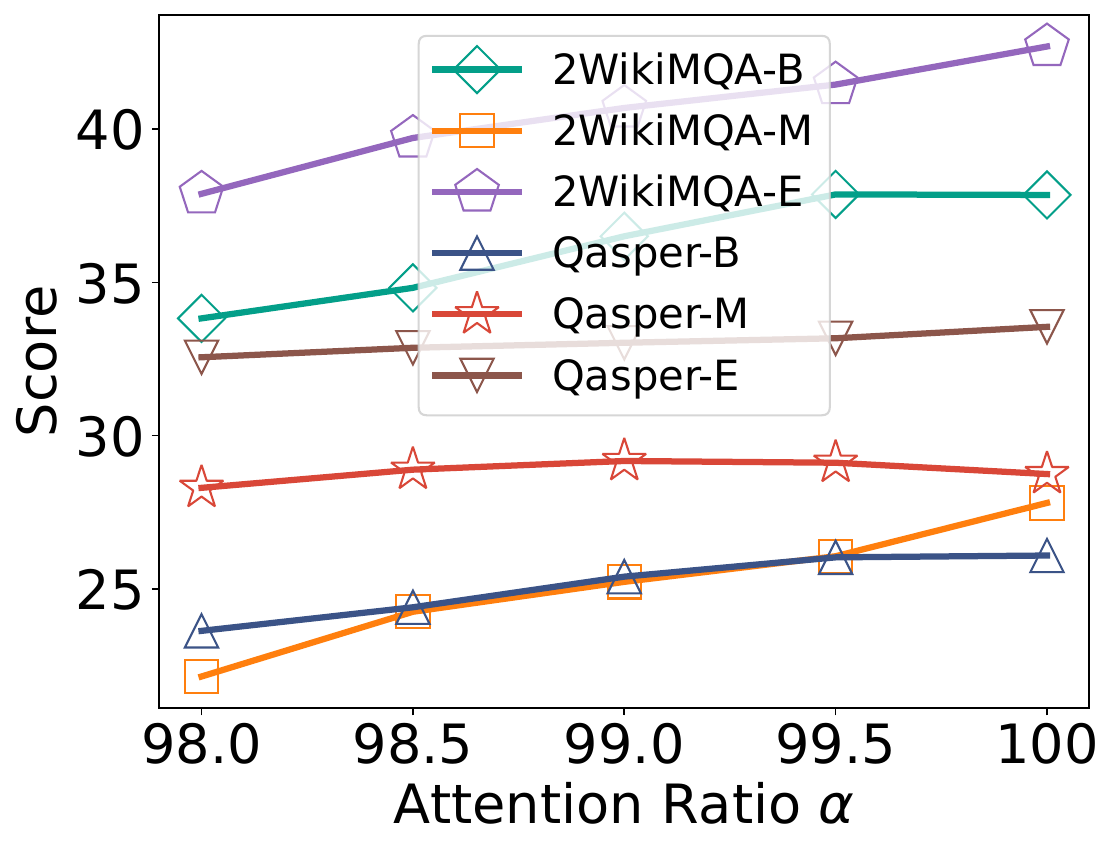}}	
	\hfill
	\subfloat[Qasper. ]		
	{\centering\includegraphics[width=0.25\linewidth]{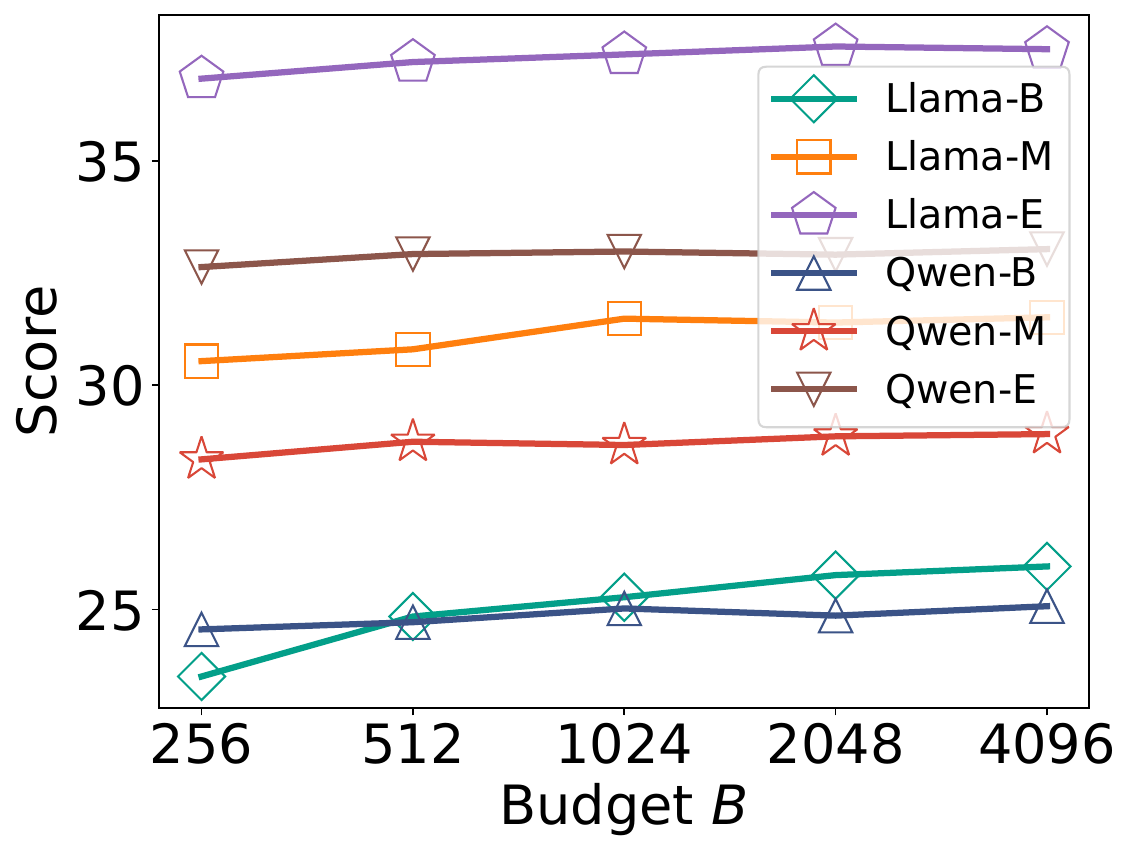}}
	\hfill
	\subfloat[Llama-3.1-8B-Instruct.]		
	{\centering\includegraphics[width=0.24\linewidth]{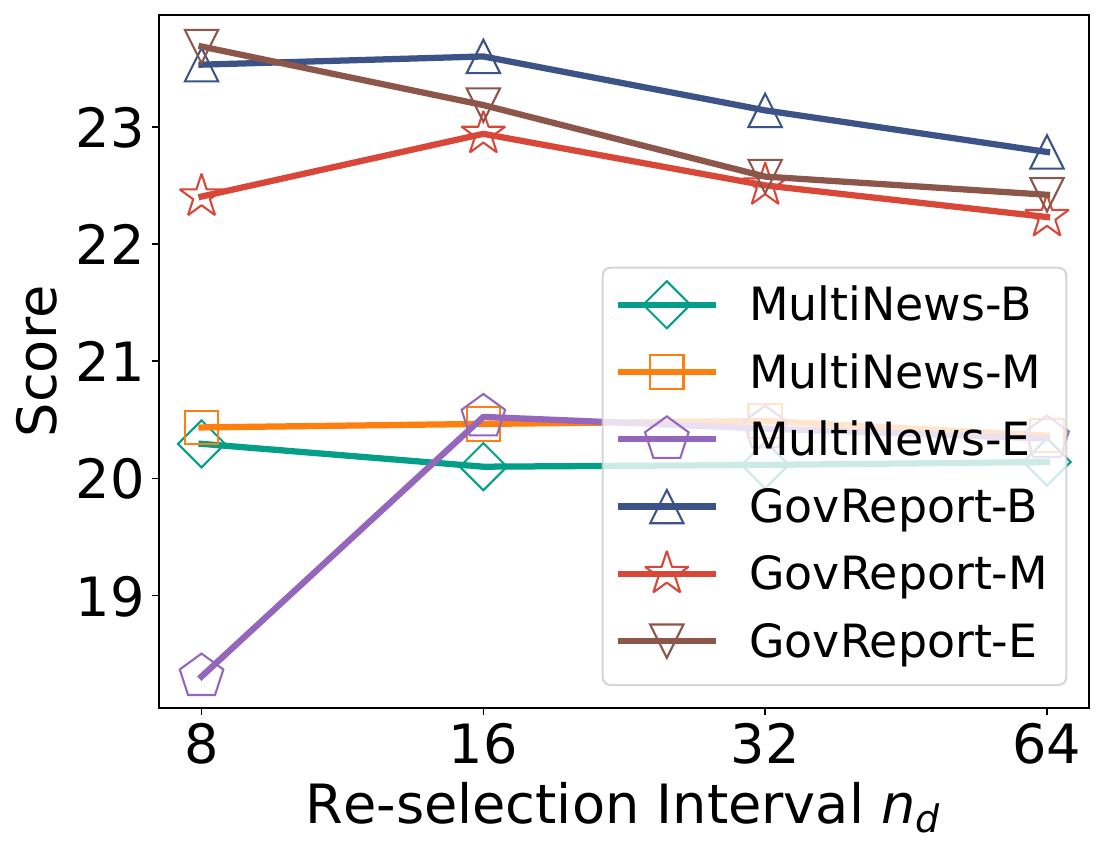}}	
	\hfill
	\subfloat[Qwen-2.5-7B-Instruct.]		
	{\centering\includegraphics[width=0.25\linewidth]{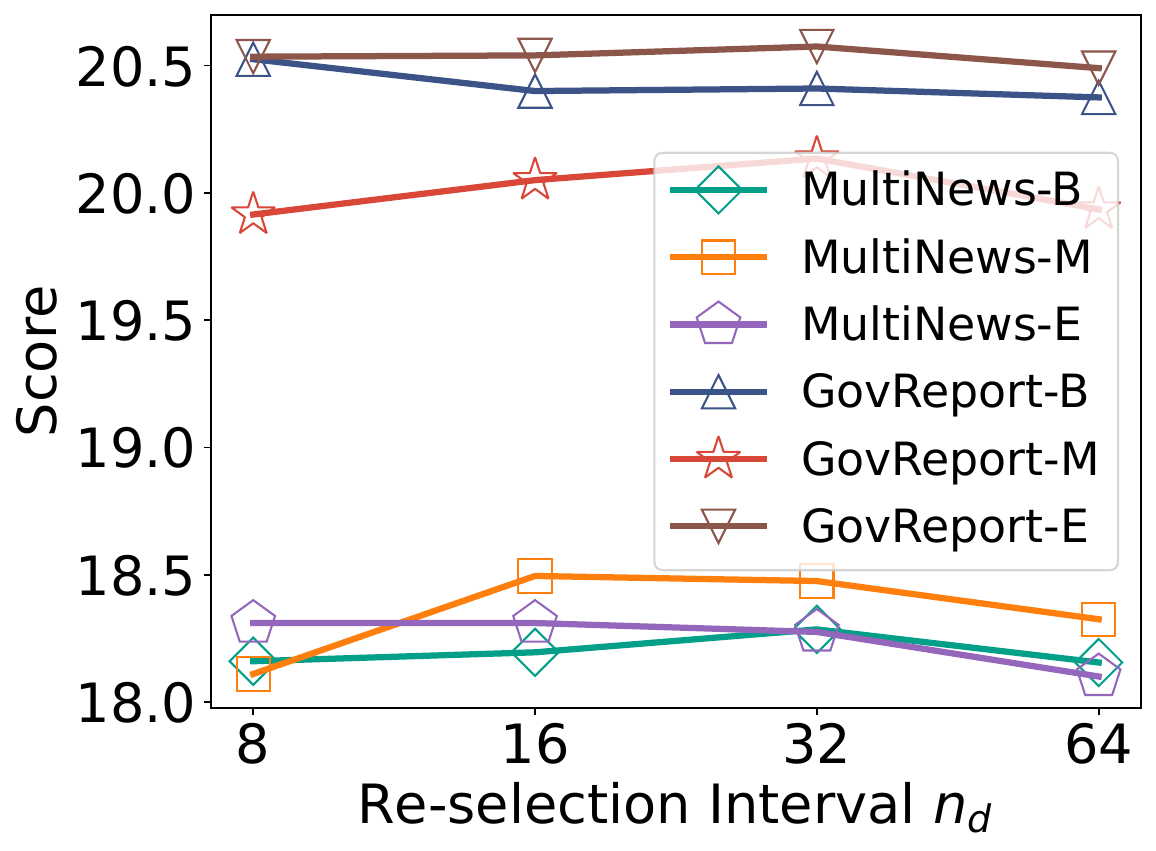}}
	\caption{Parameter sensitivity to $\alpha$ in (a), budget $B$ in (b), and decoding interval $n_d$ in (c) and (d).}
	\label{fig:param_s_B_and_param_s_nd}
\end{figure}

\subsubsection{The decoding interval  $n_d$ in Algorithm~\ref{alg:decoding}}\label{sssec:n_d_param}
The decoding interval $n_d$ controls how often LoopServe re-selects important input tokens during progressive KV compression. Smaller $n_d$ values enable frequent adaptation to changing output dependencies, improving accuracy in dynamic dialogues but increasing overhead from more KV cache updates.
We evaluate this on MultiNews and GovReport with Llama-3.1-8B-Instruct. and Qwen-2.5-7B-Instruct., covering questions at the beginning, middle, and end. As shown in Figure~\ref{fig:param_s_B_and_param_s_nd}~(c) and (d), moderate $n_d$ values (e.g., 16 or 32) strike the best balance, maintaining efficiency and robust generation quality. Very large $n_d$ reduces adaptivity, leading to lower performance on complex, multi-turn tasks.

\subsection{The Usage of LLMs for Paper Writing}
We use GPT-4o and DeepSeek-R1 to polish our paper.

\end{document}